\documentclass[acmlarge,screen]{acmart}

\usepackage{algpseudocode}
\usepackage{algorithm}
\usepackage{bm,bbm}
\usepackage{pifont}
\newcommand{\cmark}{\ding{51}}%
\newcommand{\xmark}{\ding{55}}%
\usepackage{array}
\usepackage{graphicx}
\usepackage{multirow}
\usepackage{url}
\usepackage{wrapfig}
\AtBeginDocument{%
  \providecommand\BibTeX{{%
    \normalfont B\kern-0.5em{\scshape i\kern-0.25em b}\kern-0.8em\TeX}}}

\copyrightyear{2023}
\acmYear{2023}
\setcopyright{acmlicensed}\acmConference[IoTDI '23]{International Conference
on Internet-of-Things Design and Implementation}{May 9--12, 2023}{San
Antonio, TX, USA}
\acmBooktitle{International Conference on Internet-of-Things Design and
Implementation (IoTDI '23), May 9--12, 2023, San Antonio, TX, USA}
\acmPrice{15.00}
\acmDOI{10.1145/3576842.3582372}
\acmISBN{979-8-4007-0037-8/23/05}

\newcommand{\my}{MetaMorphosis\;}

\usepackage{fancyhdr} 
\usepackage{lipsum} 

\begin{document}

\title{MetaMorphosis: Task-oriented Privacy Cognizant Feature Generation for Multi-task Learning}

\author{Md Adnan Arefeen}
\email{aa4cy@mail.umkc.edu}
\orcid{0000-0001-6486-8181}
\affiliation{%
  \institution{University Of Missouri-Kansas City}
  \city{Kansas City}
  \state{Missouri}
  \country{USA}
}

\author{Zhouyu Li}
\email{zli85@ncsu.edu}
\orcid{0000-0002-0895-4226}
\affiliation{%
  \institution{North Carolina State University}
  \city{Raleigh}
  \state{North Carolina}
  \country{USA}
}
\author{Md Yusuf Sarwar Uddin}
\email{muddin@umkc.edu}
\orcid{0000-0003-2184-0140}
\affiliation{%
  \institution{University Of Missouri-Kansas City}
  \city{Kansas City}
  \state{Missouri}
  \country{USA}
}
\author{Anupam Das}
\email{anupam.das@ncsu.edu}
\orcid{0000-0002-8961-9963}
\affiliation{%
  \institution{North Carolina State University}
  \city{Raleigh}
  \state{North Carolina}
  \country{USA}
}

\renewcommand{\shortauthors}{Arefeen et al.}
\renewcommand{\shorttitle}{MetaMorphosis}

\begin{abstract}
  With the growth of computer vision applications, deep learning, and edge computing contribute to ensuring practical collaborative intelligence (CI) by distributing the workload among edge devices and the cloud. However, running separate single-task models on edge devices is inefficient regarding the required computational resource and time. In this context, \emph{multi-task learning} allows leveraging a single deep learning model for performing multiple tasks, such as semantic segmentation and depth estimation on incoming video frames. This single processing pipeline generates common \emph{deep features} that are shared among multi-task modules. However, in a collaborative intelligence scenario, generating common deep features has two major issues. First, the deep features may inadvertently contain input information exposed to the downstream modules (violating \emph{input privacy}). Second, the generated universal features expose a piece of collective information than what is intended for a certain task, in which features for one task can be utilized to perform another task (violating \emph{task privacy}). This paper proposes a novel deep learning-based privacy-cognizant feature generation process called ``MetaMorphosis'' that limits inference capability to specific tasks at hand. To achieve this, we propose a channel \emph{squeeze-excitation} based feature metamorphosis module, \emph{Cross-SEC}, to achieve distinct attention of all tasks and a de-correlation loss function with \emph{differential-privacy} to train a deep learning model that produces distinct privacy-aware features as an output for the respective tasks. With extensive experimentation on four datasets consisting of diverse images related to scene understanding and facial attributes, we show that MetaMorphosis outperforms recent adversarial learning and universal feature generation methods by guaranteeing privacy requirements in an efficient way for image and video analytics.

\end{abstract}

\ccsdesc[500]{Computing methodologies~Computer vision tasks}
\ccsdesc[500]{Computing methodologies~Scene understanding}
\ccsdesc[500]{Security and privacy~Privacy protections}

\keywords{Multi-task learning, neural networks, collaborative intelligence, differential privacy, task privacy}



\maketitle

\section{Introduction}

\begin{figure*}
    \centering
    \includegraphics[width=0.7\linewidth]{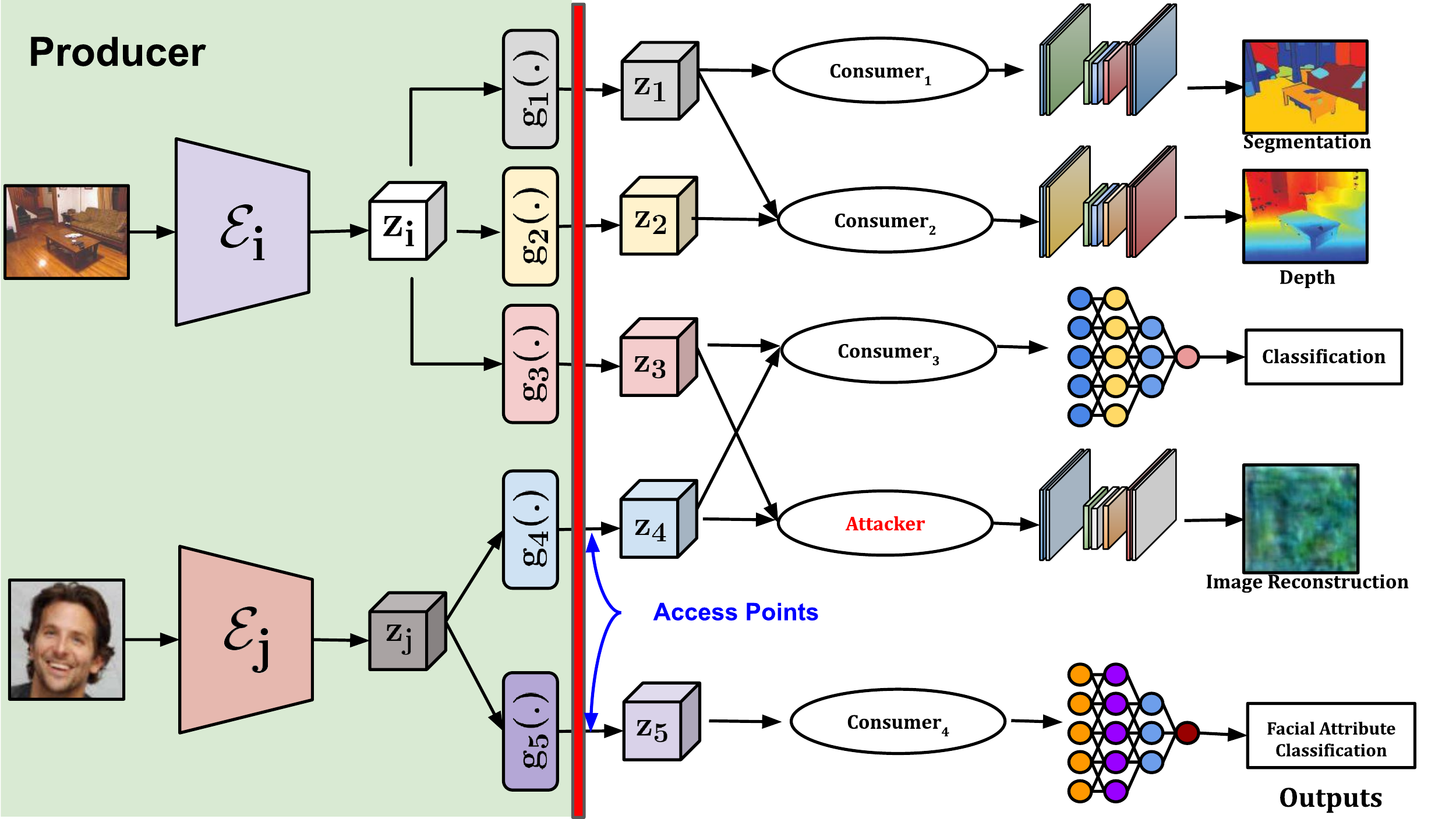}
    \caption{Overview of MetaMorphosis.}
    \label{fig:metamorph}
\end{figure*}

Computer vision-based technologies have seen widespread adoption over recent years due to improved performance. 
This use is not limited to the rapid adoption of facial recognition
technology but extends to autonomous driving~\cite{wang2022cross}, scene recognition, and more~\cite{cordts2016cityscapes,SilbermanECCV12}. 
As a result, organizations and even cities have started utilizing video feeds to carry out various automated tasks.
However, while computer vision-based technologies provide new opportunities, they also raise privacy concerns and call for novel solutions to ensure adequate privacy protection. 

One trivial way to protect sensitive information is not to send protected information outside the organization by any means, i.e., to train the deep-learning model within the respective organization providing the inputs. That implies the input-providing organizations (i.e., \emph{producers/publishers/feature providers}) also have to construct various models for different tasks, e.g., object detection, depth estimation, etc. One of the drawbacks of this approach is that organizations owning the video/input feed will also need to develop the entire analytic pipeline whose primary interest may be orthogonal to building deep learning models, such as hospitals or grocery stores. Organizations can also resort to \emph{video analytics as a service} where companies are now offering essential video processing pipelines as paid services~\cite{msweb,google}. 
However, outsourcing video feeds to cloud services also raises privacy concerns as video feeds can be used to infer various sensitive information. 
As an alternative, a hybrid approach can also be adopted where
instead of sending the raw input, some useful derived features are shared with the third party (i.e., \emph{consumers}) to prevent unintended information leakage. This approach is known as \emph{collaborative intelligence}.


In collaborative intelligence, intelligence is shared across more than one entity to split the computation overload, where one entity can run a portion of a deep model and send the intermediate partial output as ``features'' to another entity for further computation. In this way, the input can be replaced by meaningful features. 
One of the popular architectures adopted in this context is the \emph{multi-task learning} paradigm, which offers an efficient solution to reduce computational resources across different analytic tasks. The efficiency comes through the introduction of 
a shared deep layer to produce \emph{universal} features usable by all downstream tasks~\cite{li2022universal}. 
Unfortunately, it does not fully diminish the privacy concern. The shared features, also called intermediate representations, can be reverted to the actual input, thus violating \emph{input privacy} and affecting the whole notion of providing deep features rather than the input itself. 
Similarly, the universal features generated for multi-task learning, when subscribed by different downstream tasks, can also leak unintended information leading to violating \emph{task privacy}.
From a privacy and business perspective, if the task-oriented features differ, the producers can offer specific features based on consumers' objectives and hide private attribute information from each task. For example, for a given image, the feature for segmentation will be different than the feature for depth estimation or classification.

In this paper, we focus on a publisher-subscriber-based multi-party communication system where one party acts as a publisher, and the rest acts as a consumer/subscriber (Figure~\ref{fig:metamorph}). The feature publishing party also known as the feature provider or the publisher holds the data and private information, and with proper intelligent tweaking, it provides privacy-aware task-variant features to consumers. Consumers, on the other hand, consume the features rather than the input and train a task according to the features. Without loss of generality, we can assume that the producer can securely share the output with the consumer so that the consumer can train the rest of the remaining part using the task-variant features. With respect to deep learning tasks, mapping the feature to output does not violate our assumption of securing object attributes rather than the type of objects for classification cases. In the case of object detection, segmentation, or depth estimation, it can be shared to blur the attribute of the objects.

To achieve data and task privacy, we propose MetaMorphosis, which consists of two modules, (a) a \emph{private encoder} trained using differential privacy, and (b) a \emph{task metamorphosis} module for each task for task privacy. The private encoder protects identifiable information from input, which we refer to as input obfuscation. The task metamorphosis modules help to form \emph{distinct} features for each task. The privacy of the encoder depends on the requirement of privacy based on the data. So, the producer can hold private and non-private encoder submodules to offer both options to the consumers based on the privacy requirement. The whole functionality of the producer will be obscured so that the consumer cannot determine the types and characteristics of the construction of the producers.
 \begin{table*}[!htbp]
 \caption{Comparison of \my with recent literature}
\label{tab:lit-compare}
\centering
\resizebox{0.8\linewidth}{!}{
\begin{tabular}{cccccc}
\toprule
Characteristics & \my & DeepObfuscator~\cite{li2021deepobfuscator} & TIPRDC~\cite{li2020tiprdc} & ALPPTOR~\cite{xiao2020adversarial} &  P-FEAT~\cite{ding2020privacy} \\
\midrule
Input obfuscation & \cmark & \cmark & \cmark & \cmark & \cmark\\
Noisy models parameters & \cmark & \xmark & \xmark & \xmark & \xmark\\
Task privacy & \cmark & \xmark & \xmark & \xmark & \xmark \\
Scalable & Good & Poor & Poor & Poor & Poor\\  
Quantifiable privacy & \cmark & \xmark & \cmark & \xmark & \xmark \\
Feature sharing & Task-specific & Universal & Task-specific & Task-specific & Universal\\
Training budget & LOW & HIGH & HIGH & HIGH & HIGH\\

\bottomrule
\end{tabular}
}
\end{table*}

Several challenges arise when offering task-oriented privacy-aware features. Firstly, the joint training of input obfuscation and task privacy in a single phase makes the whole process challenging due to the uncertainty of leaking unintended information to task-specific features. Secondly, a sophisticated feature morphosis module is required to achieve the right balance of performance and privacy. Finally, the proposed approach has to be scalable to facilitate new tasks with minimal training effort. 
In order to address the challenges, we propose \my and specify the contribution of this work as follows. 
\begin{itemize}

 \item We propose MetaMorphosis, which ensures input obfuscation and privacy-aware task variant feature generation to prevent information leakage through the shared features while still providing acceptable outcomes for the intended tasks.
 
  
  
  \item We propose a novel task metamorphosis module ~\emph{Cross-SEC} that maintains or even improves the performance in addition to producing distinct task-specific features.
  
  \item We reduce the training time of task-specific feature generation by collaborating on the task-invariant and metamorphosis modules.
    
  \item The scope for sequential and parallel training helps \my improve scalability compared to recent adversarial learning methods, such as~\cite{li2021deepobfuscator}.

\end{itemize}

The rest of the paper is organized as follows. Section~\ref{sec:motiv} describes the motivation of our work by comparing it to similar works. Section~\ref{sec:method} explains the MetaMorphosis. We present the findings in Section~\ref{sec:results}. Section ~\ref{sec:datasets} defines the datasets and metrics that we use in the analysis. We also evaluate training and inference results after deployment of \my in Section~\ref{sec:system}. 
An ablation study is conducted to show the reasons behind choosing specific modules and parameters in Section~\ref{sec:ablation}. Related work is added in Section~\ref{sec:relatedwork}.
Finally, we conclude in Section~\ref{sec:conclusion}.



\section{Motivation and Challenges}\label{sec:motiv}
Various kinds of deep learning models~\cite{wang2019lednet, elhassan2021dsanet, chen2018encoder, seichter2021efficient, bhat2021adabins} have been proposed to resolve visual applications with multi-task learning setups such as semantic segmentation, and depth estimation, efficiently. Khattar et al.~\cite{khattar2021cross} propose a multi-task learning framework where domain-agnostic features are learned to improve the model performance on both object detection and saliency prediction tasks with limited data. Meanwhile, techniques such as knowledge distillation fit well with multi-task training where knowledge is distilled from single models by minimizing the distance, thus contributing to fast training of multi-task models~\cite{li2020knowledge,li2021universal,li2022universal}. As a universal feature is shared for all downstream tasks, it is computationally efficient but raises a privacy concern while sharing with outside agents due to offering a common feature for all tasks. Similar behavior patterns can be found in other recent literature~\cite{arefeen2021transjury,arefeen2021lightweight} where features from multiple layers of deep models are fused to form the universal features and image classification task is accomplished.

To preserve the privacy of the universal features, several adversarial learning algorithms have been proposed~\cite{li2021deepobfuscator,li2020tiprdc} to obfuscate intermediate representation. In this, adversarial decoders and classifiers are trained jointly with the intended classification task to obfuscate features~\cite{li2021deepobfuscator}. TIPRDC~\cite{li2020tiprdc} is also designed to hide private information from the feature vector while retaining the feature's utility regarding the primary task through a hybrid training algorithm. ALPPTOR~\cite{xiao2020adversarial} framework proposed a GAN loss to prevent model-inversion attack by adversarial reconstruction learning and provide task-oriented representations for binary classification tasks.
P-FEAT~\cite{ding2020privacy} proposed two adversarial objectives for privacy-preserving feature encoding-based adversarial training, which considers privacy attributes and privacy-attribute agnostic scenarios. In split federated learning~\cite{thapa2022splitfed}, intermediate features of IID data are shared with the server and the server returns the gradients back to clients. 

These methods face drawbacks at the time of adding a new task to the framework, as adversary decoders and classifiers need to be trained again with the addition of new tasks. Another disadvantage is the need for ground truth in all tasks to train the whole pipeline to prevent features from being attacked by intruders. With the development of edge computing technology, the emerging collaborative intelligent technique allows computational-constrained devices to participate as end-users where sharing of intermediate representation takes the first place to connect two entities.
Table~\ref{tab:lit-compare} compares the overview of \my with related recent literature to show the effectiveness of MetaMorphosis. Rather than training a decoder to decode the intermediate representation to a noise, \my uses differential privacy along with an intelligent split learning method, which can guarantee obfuscation of input as well as achieve target performance. \my assures \emph{task-privacy} by making the intermediate features distinct from each other, which limits the necessity to have ground truth for all tasks. With ground truth, extra DNN models are required for training an adversary classifier. At the time of addition of a new task, \my learns to make the new task features distinct from the already added tasks. Thus, \my ensures better scalability and a low training budget. To produce the distinct features, \my offers a novel metamorphosis module. 
In summary, \my answers the following key questions:
\begin{itemize}
\item How to reduce the input information leak while sending deep features rather than the input itself?
\item How to overcome privacy issues  regarding universal features for all tasks? 
\item How to design a lightweight task metamorphosis module
so that the performance drop should be negligible and almost similar to the performance of a single task?
\end{itemize}

\section{\my}\label{sec:method}
Generating features for different tasks is the core part of \my and as a result, several considerations are undertaken in the construction of the producer to enhance the target performance and privacy, reduce memory issues, and latency of the system.

\subsection{Privacy Cognizant Feature Generation}
 At first, task-oriented single models cannot be provided due to zero task privacy for independent training. In addition, memory requirements will increase when new tasks are assigned. So, a multi-task model is required to reduce the number of models. In this way, one model can provide 
a universal encoder to produce the features for all tasks. But the drawback of the latter method for producing universal representation lies in the degradation of performance and privacy in some cases for de-correlated tasks. For example, the data owner can issue a restriction on reconstructing the images from the encoded features for facial image attribute classification. But for semantic segmentation or depth estimation tasks, privacy can be imposed on the feature generation so that unique features are generated for each task at hand. A universal representation fails to either provide high accuracy for all tasks or prevent privacy attacks due to providing the same features for all tasks, e.g., the same features are provided for gender classification and smile classification from facial images.

Furthermore, a producer cannot offer any arbitrary feature for any task. To claim a good performance over some offered tasks, it needs to train the whole pipeline in an end-to-end fashion to provide a meaningful feature for a certain task. The notations used throughout the paper to describe \my are shown in Table~\ref{tab:notation}.

To construct the model in a cost-effective fashion and to reduce the model size as well, we divide the producer into a feature extractor part (encoder $\mathcal{E}$), a \my module ($\mathbf{g(.)}$), and the target task. For clarity, we use the producer, feature extractor, and encoder as the same entity, $\mathcal{E}$ or $\mathcal{E}_p$ (encoder trained with differential privacy) throughout the paper. To produce meaningful features, the producer goes through a full training effort respecting the input obfuscation and task privacy. After training the whole model, the producer splits the model into two parts: one part includes a semi-universal encoder for some sub-tasks and unique transformer modules for each of the tasks. For other similar sub-tasks, another similar feature extractor module may exist. The remaining part will be hidden from the outside environment and is kept on the producer side only. Thus, the producer will offer access points for only subscribed consumers for the respective tasks. But where to split is an issue in maintaining communication vs. computation trade-off~(see Sections~\ref{sec:system} and \ref{sec:ablation}). Although the earlier layers are suitable for a lightweight encoder, they may be prone to reconstruction image attack very easily. It is difficult to retain the original image from the layers closer to the outputs. The feature provider should also offer features so that consumers will produce the final output for a task with minimal effort.

\begin{table}[!htpb]
    \centering
    \caption{Notation}
    \label{tab:notation}
    \resizebox{0.9\columnwidth}{!} {%
    \begin{tabular}{cc|cc}
    \toprule
         Description & Notation & Description & Notation  \\
         \midrule
         Input & $\mathbf{x}$ & Output & $\mathbf{y}$\\
         Producer Model & $\mathcal{M}_f$ & Consumer Model & $\mathcal{M}_c$ \\
         Task & $\mathbf{T}$ & Task features & $\mathbf{z}_{1,2,...,\mathbf{T}}$\\
         Task Metamorphosis Module & $\mathbf{g}$ & Encoder & $\mathcal{E}$ \\
         Private Encoder & $\mathcal{E}_p$ & Private Feature & $\mathbf{z}_p$\\
         MetaMorphosis & $\mathcal{G}$ & Decoder & $\mathcal{E}^{-1}$\\
         \bottomrule
    \end{tabular}}
\end{table}

To design MetaMorphosis, suppose a model $\mathcal{M}_f$ is trained by the producer that provides features for a corresponding task $\textbf{T}_i$ for an input $\mathbf{x}$. At the time of inference, the producer will share the intermediate features as output, denoted by $\mathbf{z}_i$, from a portion of the model $\mathcal{G}_i$ where $\mathcal{G}_i \in \mathcal{M}_f$. After getting the features $\mathbf{z}_i$ instead of the raw input $\mathbf{x}$, the consumer runs its own model $\mathcal{M}_c$ on $\mathbf{z}_i$ and produces $\mathbf{\hat{y}}_i$ for task $\mathbf{T}_i$ where the ground truth is $\mathbf{y}_i$.
Mathematically, it can be written as follows.
\begin{equation}\label{eqn-1}
\begin{aligned}
\min \mathbf{y}_i \sim \mathbf{\hat{y}}_i &= \mathcal{M}_c \circ (\mathcal{G}_i \circ \mathbf{x})  = \mathcal{M}_c (\mathbf{z}_i)\\
\textrm{s.t.} \quad & Acc(\mathbf{T}_i|\mathbf{z}_i) - \sum_{i \neq j}Acc(\mathbf{T}_j|\mathbf{z}_i) \approx Acc(\mathbf{T}_i|\mathbf{z}_i) \\
 & Acc(\mathbf{T}_i|\mathbf{z}_i) \geq \xi \quad; \quad 0 < \xi < 1\\
\end{aligned}
\end{equation}
In Equation~\ref{eqn-1}, the objective is to maintain the target performance ($Acc$) for task $\textbf{T}_i$ and obfuscate the input $(\mathbf{x})$ and features $(\mathbf{z})$ to limit the accuracy of all other tasks with the current task features. As $\mathcal{M}_c$ and $\mathcal{G}_i$ will not be processed by the same party, a few privacy considerations should be established. Based on this, we can divide the overall producer construction into two components: (1) \textbf{Input obfuscation}, and (2) \textbf{Task-privacy}.\\ In the next subsections, we will investigate thoroughly Equation~\ref{eqn-1} in terms of input obfuscation and task privacy and discuss the final equation as shown in Equation~\ref{eqn-8}.

\subsection{Input Obfuscation} 
By input obfuscation, we mean the input should be made private so that the features provided by the producer cannot be converted back to the original input by an attacker. If $z = \mathbf{\mathcal{E}(x}; \theta_\mathbf{\mathcal{E}})$, then it is nearly impossible to find a $\mathbf{\mathcal{E}}^{-1}$ so that $\mathbf{\mathcal{E}}^{-1}(\textbf{z}) = \textbf{x}$.\\
To ensure input obfuscation, we propose an efficient use of differential privacy which is defined as follows.

\begin{definition}\label{dp-def}
    If $d$ and $d'$ are two adjacent inputs of $D$ that differ by at least one sample and they follow a certain condition such that
    \begin{equation}\label{eqn:dp}
        Pr[\mathbf{f}(d) \in D] \leq e^\epsilon Pr[\mathbf{f}(d') \in D] + \delta
    \end{equation}
    where, $\mathbf{f}$ is a randomized function, i.e., $\mathbf{f} : D \xrightarrow{} \mathcal{R}$, then $\mathbf{f}$ satisfies $(\epsilon,\delta)$ differential privacy (DP)~\cite{abadi2016deep}.
\end{definition}

Definition~\ref{dp-def} is also known as R$\acute{e}$nyi-differential privacy~\cite{mironov2017renyi} which is a relaxed version of $\epsilon$-DP with a $\delta$. From Equation~\ref{eqn:dp}, we see that the higher the value of $\epsilon$, the lower the privacy bound.
Differential privacy operations in deep learning models are shown in Algorithm~\ref{algo-dp} where noises are added with gradients before updating the parameters~\cite{abadi2016deep}. To get a desired $\epsilon$, the noise $\sigma$ can be chosen for a number of training steps $T$, batch size $q$, and $\epsilon < c_1 q^2 T$ as the following Equation~\ref{eqn-3}~\cite{abadi2016deep}. Here, $c_1$ and $c_2$ are constants.
\begin{equation}\label{eqn-3}
    \sigma \geq c_2 \frac{q\sqrt{T\log \frac{1}{\delta}}}{\epsilon} 
\end{equation}
\begin{algorithm}[!htpb]
  \caption{Differential Privacy Operations}\label{algo-dp}
  \begin{algorithmic}[1]
    \Function{Gradient Computation}{.}
    \State $g_t (z_i) \xleftarrow{} \nabla_{\theta_t} \mathcal{L} (\theta_t , z_i) $
    \EndFunction
    
     \Function{Gradient Clipping}{.}
    \State $\Bar{g_t}(z_i) \xleftarrow{} g_t (z_i) / {\max(1,\frac{||g_t (z_i)||_2}{C})} $\Comment{Gradient Clipping with certain threshold $C$}
    
    \EndFunction
    \Function{Noise Addition}{.}
    \State $g_t (z_i) \xleftarrow{} \frac{1}{n} \sum_i (\Bar{g}(z_i) + \mathcal{N}(0,\sigma^2C^2\mathbf{I}) $ \Comment{Adding noise to the gradient}
    
    \EndFunction
  \end{algorithmic}
\end{algorithm}
Rather than adding differential privacy in the input as shown in recent literature~\cite{li2021deepobfuscator}, we perform DP in model parameters for input obfuscation. As in MetaMorphosis, the producer holds the feature extractor part only, to generate meaningful features, the feature extractors along with target classifiers are required to train jointly. A split learning method can reflect the scenario where a model is split into the feature extractor part and the classifier part. So, we propose differential privacy with split learning to achieve input obfuscation.\\
\textbf{Split learning with differential privacy:} 
Input obfuscation results in a trade-off between utility vs. privacy. 
As the main goal of \my is to provide task-specific features, injecting noises to ensure DP into the whole model parameters while training to ensure only input obfuscation to the encoder $\mathcal{E}_p$ in unnecessary and it affects the task performance. In most cases, the consumer resides in the public domain and making the consumer model parameters private have little effect on overall privacy constraints. As a result, making only producer model parameters private will suffice our goal. In that case, during training of the feature generator, the provider uses differential privacy to ensure the input obfuscation of the generator only while learning the intended task using the split learning method~\cite{thapa2022splitfed}. A detailed discussion of utility vs. privacy is discussed in Section~\ref{sec:exp-res}.

\subsection{Task Privacy}
As we consider the service provider (feature-provider) as an MLaaS (Machine Learning as a Service) platform, the service provider/producer will offer meaningful features for certain tasks to the public domain. In this case, instead of providing a single universal interface, the service provider offers multiple access points for some task-privacy-related features to the subscribers/consumers. By task privacy, we mean the features used for one task will not perform well for another task. Mathematically, the deep features generated for one task should be far apart from another task. 
\begin{figure}
    \centering
    \includegraphics[width=0.7\columnwidth]{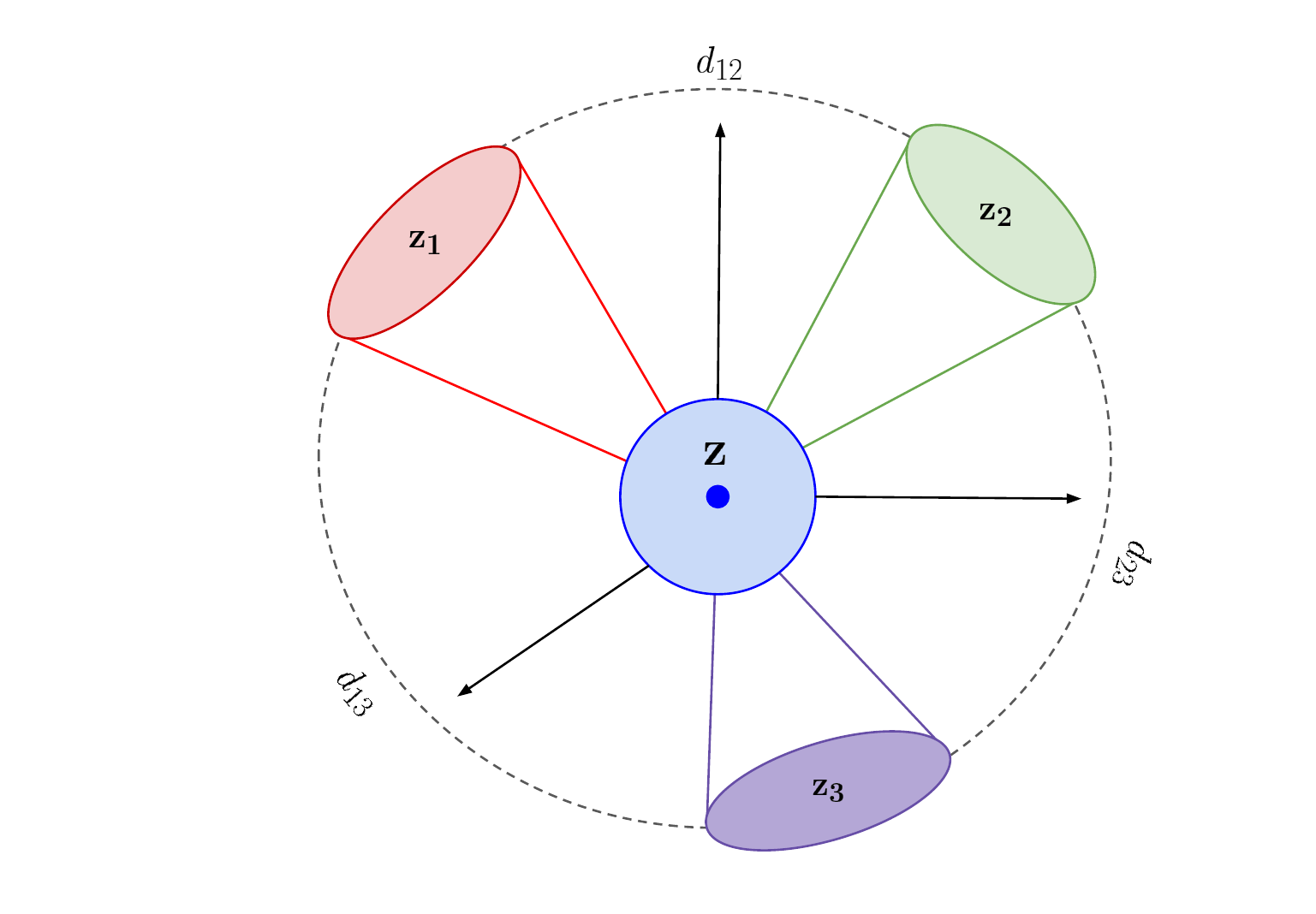}
    \caption{Pictorial representation of privacy-aware task feature generation.}
    \label{fig:model-tp}
\end{figure}
We formally formulate task privacy as follows. 
For any input $x$, if there exist $n$ feature extractors $(\mathcal{G}_{1...n})$ for multi-task learning, then the optimal similarity between any two feature extractors $\mathcal{G}_i(.)$ and $\mathcal{G}_j(.)$ satisfies the following equation.

\begin{equation}
   \sum_{i\neq j}^n\mathcal{S}(\mathcal{G}_i(\mathbf{x}),\mathcal{G}_j(\mathbf{x}))\approx 0
\end{equation}

\noindent
where $\mathcal{S}(., .)$ denotes a similarity function between two features. To ensure \emph{task-privacy}, the summation of the similarity between features will be theoretically 0. A pictorial representation of task-privacy transformation on certain feature $\textbf{z}$ is shown in Figure~\ref{fig:model-tp}.

\textbf{Task Metamorphosis Module: }
In MetaMorphosis, we propose a novel feature module for each task, instead of sharing a common feature for all consumers. The main goal of each metamorphosis module is to produce task-specific features as distinctively as possible with an assurance of better performance for the respective task. In this way, the attacker is unable to produce meaningful features for the specific private task. To ensure better performance, the metamorphosis module should capture the most informative feature of the task. To achieve this, we propose an attention-based metamorphosis module, Cross-SEC, that enables the general features $\mathcal{E}$ to be more informative. At first, we split the $\mathcal{E}(x)$ into $k$ splits. For each split, we get the global attention of the features. Using the Conv-Linear-ReLU-Linear module, we transform the features and add a Sigmoid activation layer to get the attention values. Then, the attention is swapped between the splits following a Conv ($1\times 1$) layer. At the time of joint training of $\mathcal{E}$ and $\mathbf{g}(.)$, the swapping of attention values will try to make $\mathcal{E}$ more informative as it avoids making only some channels of $\mathcal{E}$ more informative.   

\begin{figure}[!t]
    \centering
    \includegraphics[width=0.8\columnwidth]{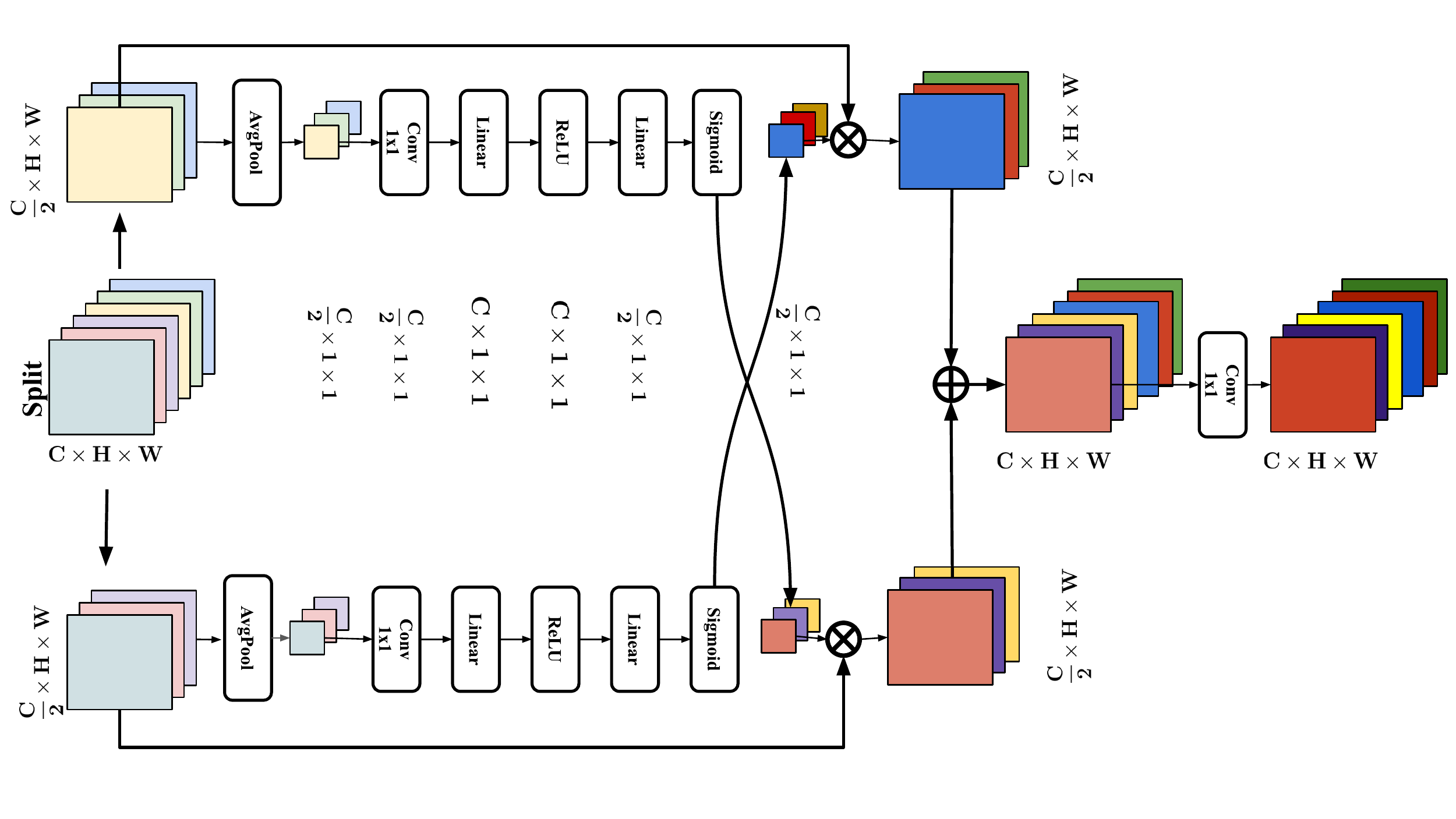}
    \caption{Cross-SEC Metamorphosis Module}
    \label{fig:cross-sec}
\end{figure}

The metamorphosis module is shown in Figure~\ref{fig:cross-sec} with the shape for each layer. To make the task features distinct, we use a similarity metric as used in recent literature~\cite{arefeen2021transjury,li2021deepobfuscator}. In this case, we use the SSIM metric to compare the structural similarity among task features, and in the loss function, it learns to project them far from each other based on the weight given to this metric. The task-privacy loss function can be written as follows.
\begin{equation}
    \ell_{tp} = \sum_{i,j \in \mathbf{T}} \mathbbm{1}[\mathbf{T}_i \neq \mathbf{T}_j] \, \mathcal{S} \Bigl( \mathbf{g}_i (\mathbf{\mathcal{E}(x)}), \mathbf{g}_j(\mathbf{\mathcal{E}(x)}) \Bigr)
\end{equation}

This metric will be added to the loss function with other task performance losses to achieve the desired behavior of MetaMorphosis. Together, we can write the whole equation as follows.
\begin{equation}
\begin{aligned}
loss = \sum_{i=1}^{\mathbf{|T|}} \mathcal{L}_i + \omega\sum_{i,j \in \mathbf{T}} \mathbbm{1}[\mathbf{T}_i \neq \mathbf{T}_j] \, SSIM \Bigl( \mathbf{g}_i (\mathbf{\mathcal{E}(x)}), \mathbf{g}_j(\mathbf{\mathcal{E}(x)}) \Bigr)
\end{aligned}
\end{equation}
Here, $\omega$ controls the weight of the distance loss function to overall loss. To make the feature generator more efficient, we can use a single encoder and multi-task transformer modules for a group of tasks. To assure task privacy and input obfuscation, we can rewrite the function $\mathcal{G}$ as a composition of private-encoder $\mathcal{E}_p$ that prevents exposing the private information and a task transformer module that converts the $\mathbf{g}$. For task privacy only, the encoder can be non-private ($\mathcal{E}$). 
\begin{equation}
    \mathcal{G} (\mathbf{x}) = (\mathbf{g} \circ \mathcal{E}_p) (\mathbf{x})
\end{equation}
We can combine these two aspects of privacy and elaborate the Equation~\ref{eqn-1} and relax the constraints to achieve efficient training as follows.
\begin{equation}
\begin{aligned}
\min \mathbf{y}_i \sim \mathbf{\hat{y}}_i &= (\mathcal{M}_c \circ \mathcal{G}_i)(\mathbf{x}) 
           = (\mathcal{M}_{c} \circ \mathbf{g}_i \circ \mathcal{E}_p) (\mathbf{x})
           = \mathcal{M}_c (\mathbf{z}_i)\\
 \textrm{s.t.} \quad & \mathbf{\mathcal{E}_p}^{-1}(\mathbf{z}_i) \neq \mathbf{x}\\
  & \sum_{i\neq j}^\mathbf{T}\mathcal{S}(\mathcal{G}_i(\mathbf{x}),\mathcal{G}_j(\mathbf{x}))\approx 0    \\
\end{aligned}
\label{eqn-8}
\end{equation}

\subsection{\my Training Scheme}
\begin{figure}[!htbp]
    \centering
    \includegraphics[width=\columnwidth]{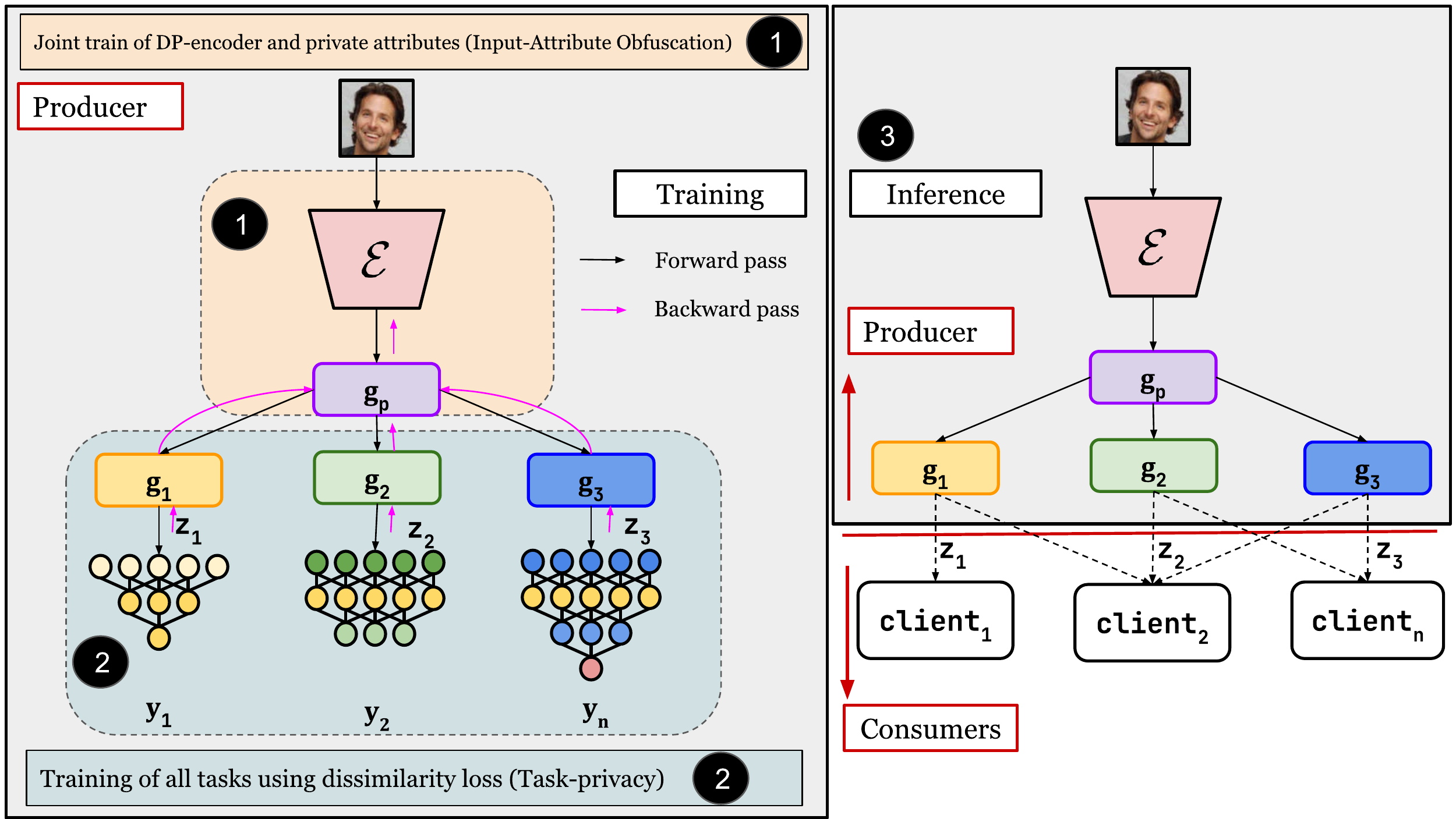}
    \caption{Producer training and inference scheme}
    \label{fig:producer-training}
\end{figure}

The training scheme of \my is shown in Algorithm~\ref{algo-meta}. \my obfuscates the input and the tasks in two phases. If input obfuscation and private attribute obfuscation are imposed, then the encoder with the privacy attribute classifier is trained jointly at Phase 1 [line 9 in Algorithm~\ref{algo-meta}]. After the completion of Phase 1, in Phase 2, the task variant metamorphosis modules are trained along with the respective classifiers [line 10 in Algorithm~\ref{algo-meta}], where the encoder trained from phase 1 is kept fixed to provide features. In line 9, $\mathcal{M}_p$ refers to the private classifier (i.e. gender for face images) that the publisher intends to hide. It will train other tasks i.e. $\mathcal{M}_{c_i}$ by hiding private information using task privacy [line 6-7, 10 in Algorithm~\ref{algo-meta}]. After the completion of two-phase training, the task features are ready for the consumers. Figure~\ref{fig:producer-training} shows the steps of the training and inference scheme of MetaMorphosis. 
At the time of inference, the producer will offer access points ($\mathbf{z_{1,2,...,\mathbf{T}}}$) for different tasks.
\begin{algorithm}
  \caption{\my}\label{algo-meta}
  \begin{algorithmic}[1]
    \If{Input obfuscation only}
        \State $\mathbf{\hat{y}}_{c_i} = \mathcal{M}_{c_i} \circ \mathcal{E}_p \circ \mathbf{g}_{i} (\mathbf{x})$ \Comment{forward pass}
        \State Compute $\mathcal{L}(\mathbf{y}_{c_i},\mathbf{\hat{y}}_{c_i})$ 
        \State Update $\theta_{\mathbf{g}_i}, \theta_{\mathcal{E}_p}$ using Algorithm~\ref{algo-dp},Update $\theta_{c_i}$ \Comment{backward pass}
    \ElsIf{Task-privacy only}
    \State Compute $\sum_{i=1}^{\mathbf{|T|}} \mathcal{L}_i +  \omega\sum_{i\neq j} SSIM(\mathbf{g}_i(\mathbf{z}),\mathbf{g}_j(\mathbf{z}))$
    \State Update $\theta_{\mathbf{g}_i}, \theta_{\mathcal{E}} ,\theta_{c_i} \forall i\in \mathbf{T}$
  \Else
    \State At phase 1, do steps 2-4 to joint train the $\mathcal{E}_p$ and $\mathcal{M}_p$
    \State At phase 2, using 6-7 train $\mathbf{g}_i$, $\mathcal{E}_p$, and  $\mathcal{M}_{c_i}$
   \EndIf
  \end{algorithmic}
\end{algorithm}
\subsection{Threat Model}
Before going into detail on experimental results, we describe the attacker model in this section. For input obfuscation, we extract the private encoder features and use a decoder model (Figure~\ref{fig:dp-effective}) to act as an attacker trying to reconstruct the image. For task privacy, we assume the consumer portion of the model architecture is as same as the producer model architecture while training. In this, for all cases of classifiers, we use the same model architecture (ResNet-18) to act as an attacker. At the time of task privacy evaluation, we interchange the task metamorphosis module but keep the classifier layers and weights intact as the producer for the attacker.
In Section~\ref{sec:results}, we implement and evaluate \my on different types of tasks and compare \my with recent relevant literature.
\section{Evaluation}\label{sec:results}
\subsection{Datasets and Metrics}\label{sec:datasets}
To have a deep understanding of \my performance, we evaluate the task-privacy algorithm in different domains with various task complexities. To simulate an indoor robot scenario, we use the NYU-v2 dataset~\cite{SilbermanECCV12}, which contains 1449 indoor images with ground truth images on three tasks, i.e., semantic segmentation, depth estimation, and surface normal estimation. We have used 795 images for training \my and evaluated the rest 654 images. To simulate the road scene-based tasks, we use the CityScapes dataset~\cite{cordts2016cityscapes}, which contains 3475 vehicle road scene views. Based on recent literature, we use the 2975 images for training and the rest 500 images for testing the performance of MetaMorphosis. We also use a large facial attribute dataset named CelebA~\cite{liu2015faceattributes} that includes more than 200000 images (162000 for training, 40000 for testing) to show multi-binary classification-based task privacy. For the multi-class classification scenario, we use StateFarm (a total of 22424 images, use 17934 for training and 4490 for testing ) to validate the input obfuscation and task privacy. 

\subsection{Implementation details}\label{sec:im-details}
We use PyTorch to implement \my and to execute the training we use $4\times$ 16 GB NVIDIA RTX A4000 workstation for all datasets. We use cross-entropy loss as shown below for segmentation and compute the mean Intersection over Union (mIoU), and pixel accuracy as a performance metric as referred to~\cite{li2020knowledge}.
\begin{equation}
    \mathcal{L}_{seg} = -\frac{1}{m} \sum_{i=1}^{m} y_i \log \hat{y}_i +  (1 - {y}_i) \log (1 -\hat{y}_i)
\end{equation}
For depth estimation, we use the absolute error as described by~\cite{li2020knowledge} and also use it in the loss function to minimize the depth error.
\begin{equation}
    MAE_{depth} = \frac{1}{N} \frac{\sum_{i,j} |y_{i,j} - \hat{y}_{i,j}|}{\sum_{i,j} \mathbbm{1} [y_{i,j}>0]}
\end{equation}
In surface normal estimation, we use the mean and median of per pixel error and compute the fraction of error within a certain threshold (11.25, 22.5, 30). For Cityscapes and NYU-v2, we use Adam optimizer with an initial learning rate (LR) $1\times10^{-4}$. A step learning rate scheduler changes the LR with step size 100 and $\gamma=0.5$. We ran each experiment for 200 epochs and chose the best model with the smallest average training error of all tasks. We then use the best model for prediction. For NYU-v2, we use 13 classes for segmentation, and for CityScapes, we use seven classes. The batch size is 8 and 2 for Cityscapes and NYU-v2, respectively. We use 0.001 as weight on SSIM loss while adding task privacy. For the CelebA and StateFarm datasets, we use $\epsilon=4$, and 1.2 as the maximum gradient clipping ($C$) for both StateFarm, and CelebA, and $\delta=10^{-5},10^{-6}$ respectively. The batch size is set to 64, and AdamW~\cite{loshchilov2017decoupled} optimizer with LR=$10^{-4}$. For data transformation, we resize to make the images to $64\times64$ pixels, and use RandomHorizontalFlip at training. The normalization parameters are used as same as ImageNet. We use Opacus~\cite{yousefpour2021opacus} to train the model using differential privacy. We have chosen the lightweight ResNet-18 model, split it in half at different points, and used the first portion as the encoder and the rest as the private and intended classifier.

\subsection{Experimental Results}\label{sec:exp-res}
\textbf{CityScapes and NYU-v2:} As \my imposes privacy constraints either on the content or on the task or on both. Considering task privacy we focus on NYU-v2 and Cityscapes dataset. For both datasets, we use SegNet model~\cite{li2020knowledge} with knowledge distillation during training. Table~\ref{tab:res-city} shows the results on the test set using KD-MTL~\cite{li2020knowledge} where privacy-aware feature generation is absent. With the addition of cross-SEC metamorphosis module and SSIM loss function, we compare the utility as the performance metric for both and compare task privacy based on the interchange of the module. For having the distilled knowledge, we first train every single model to train a single task. Then using Algorithm~\ref{algo-meta} for task-privacy only, we train the joint model to produce output similar to every single model and add the privacy loss to make each task features distinct. We joint train the segmentation, depth, and surface-normal estimation for NYU-v2 using task-transformer module and compare it without the task-transformer module and without task privacy. For segmentation results, we observe a $7.61\%$ higher mean Intersection Over Union (mIOU) than KD-MTL and a $2.25\%$ higher pixel accuracy metric. Compared to the depth estimation results, \my achieves almost the same results for absolute error and a little worse in relative error. having a $conv (1\times 1)$ for each task. As cross-SEC transformer generalizes better  task features. For StateFarm dataset, we train for 20 epochs. For CelebA we train for 10 epochs.

\begin{table}[!t]
\caption{Test set results on CityScapes~\cite{cordts2016cityscapes} dataset. \my achieves higher pixel accuracy for segmentation and almost the same absolute error for depth.}
\label{tab:res-city}
\centering
\resizebox{\columnwidth}{!} {%
\begin{tabular}{cccc|cc}

\toprule[1.2pt]
Model & Size  & \multicolumn{2}{c|}{Segmentation} & \multicolumn{2}{c}{Depth} \\

 &(MB) & mIoU $(\uparrow)$ & Pix Acc $(\uparrow)$ & Abs Err $(\downarrow)$ & Rel Err $(\downarrow)$ \\
\midrule
KD-MTL~\cite{li2020knowledge}   & 300.90  & 52.18 & 91.24 & 0.0140 & 28.90 \\
\my  & 307.00 & \textbf{59.79} & \textbf{93.49} & 0.0141 & 31.89\\
\bottomrule[1.2pt]
\end{tabular}
}
\end{table}

To show the task-privacy evaluation, we use the trained segmentation Cross-SEC module features to infer segmentation and depth estimation and vice versa for depth estimation. From Table~\ref{tab:tp-eval}, we observe a sharp drop in the performance of both tasks. For segmentation, the mIoU and pixel accuracy drop down to $1.47\%$ and $7.33\%$, respectively. For depth estimation, the absolute error is almost $10\times$ higher using the segmentation feature. 

\begin{table}[!t]
\caption{Task-privacy evaluation of Cityscapes~\cite{cordts2016cityscapes}. Use of one task metamorphosis module to evaluate the performance of other tasks. Using depth features for segmentation, lower mIoU, and pixel accuracy indicate higher task privacy and vice versa. For depth estimation, the higher error with segmentation features indicates higher task privacy and vice versa.}
\label{tab:tp-eval}
\centering
\resizebox{\columnwidth}{!}{
\begin{tabular}{cccc|cc}

\toprule[1.2pt]
Metamorphosis & Methods & \multicolumn{2}{c|}{Segmentation} & \multicolumn{2}{c}{Depth} \\
 Module (Replaced) &  & mIoU $(\uparrow)$ & Pix Acc $(\uparrow)$ & Abs Err $(\downarrow)$ & Rel Err $(\downarrow)$ \\
\midrule

 --- &  \my & 59.79 & 93.49 & 0.0141 & 31.89\\
\midrule
Segmentation & \my & 59.79 & 93.49 & {\color[HTML]{FE0000}\underline{0.1079}} & {\color[HTML]{FE0000}\underline{99.07}}\\
Depth &  \my & {\color[HTML]{FE0000}\underline{1.47}} & {\color[HTML]{FE0000}\underline{7.33}} & 0.0141 & 31.89\\
\bottomrule[1.2pt]
\end{tabular}
}%
\end{table}
\begin{figure}[!htpb]
\centering
\includegraphics[width=\columnwidth]{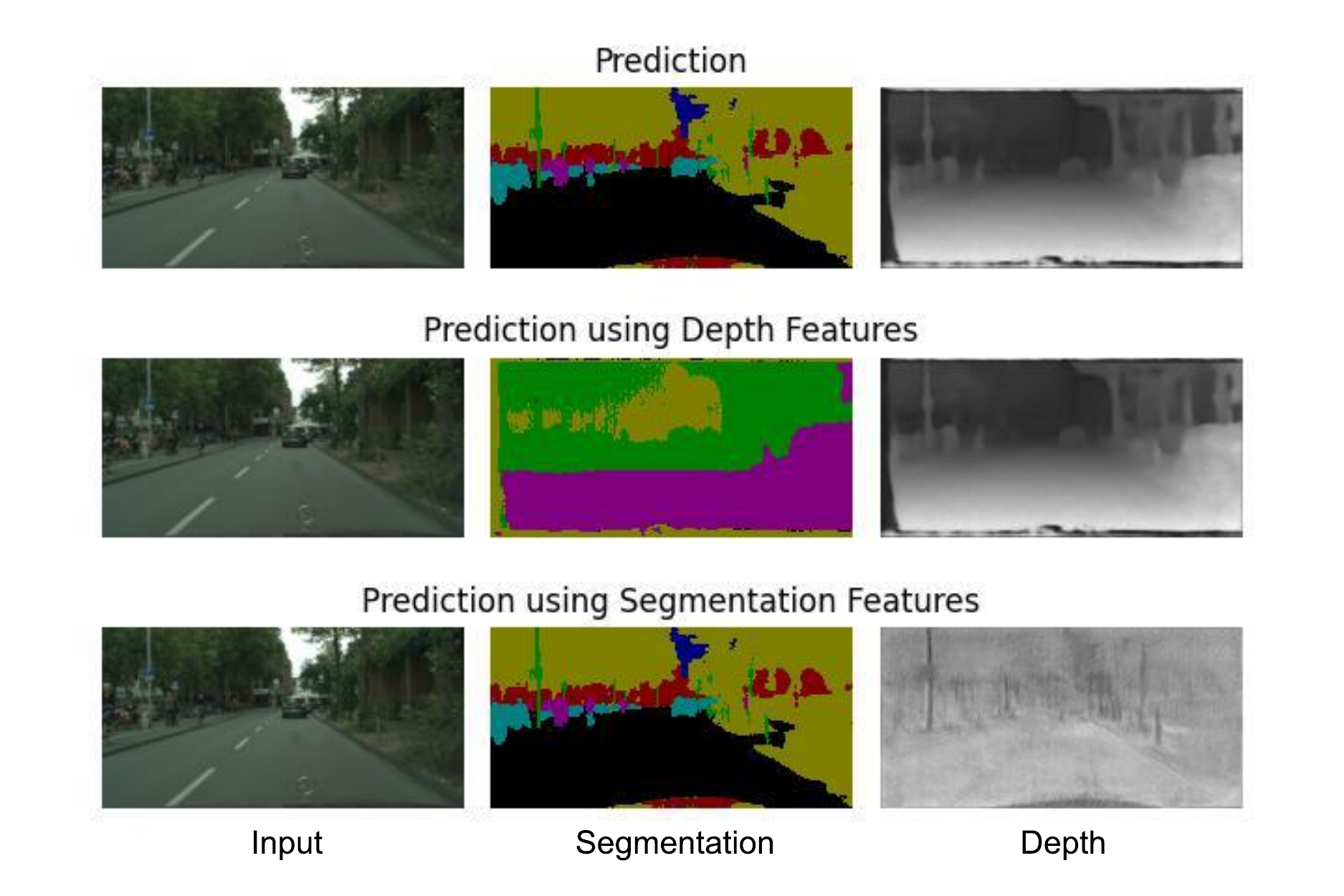}
    \caption{Qualitative analysis of task privacy. Prediction results of the segmentation and depth estimations of the CityScapes dataset using \my(first row). The task metamorphosis modules are interchanged and the output is produced (second and third row).}
    \label{fig:quant-cs}
\end{figure}

\begin{figure}
    \centering
    \begin{minipage}{0.45\textwidth}
        \centering
        \includegraphics[width=0.9\linewidth]{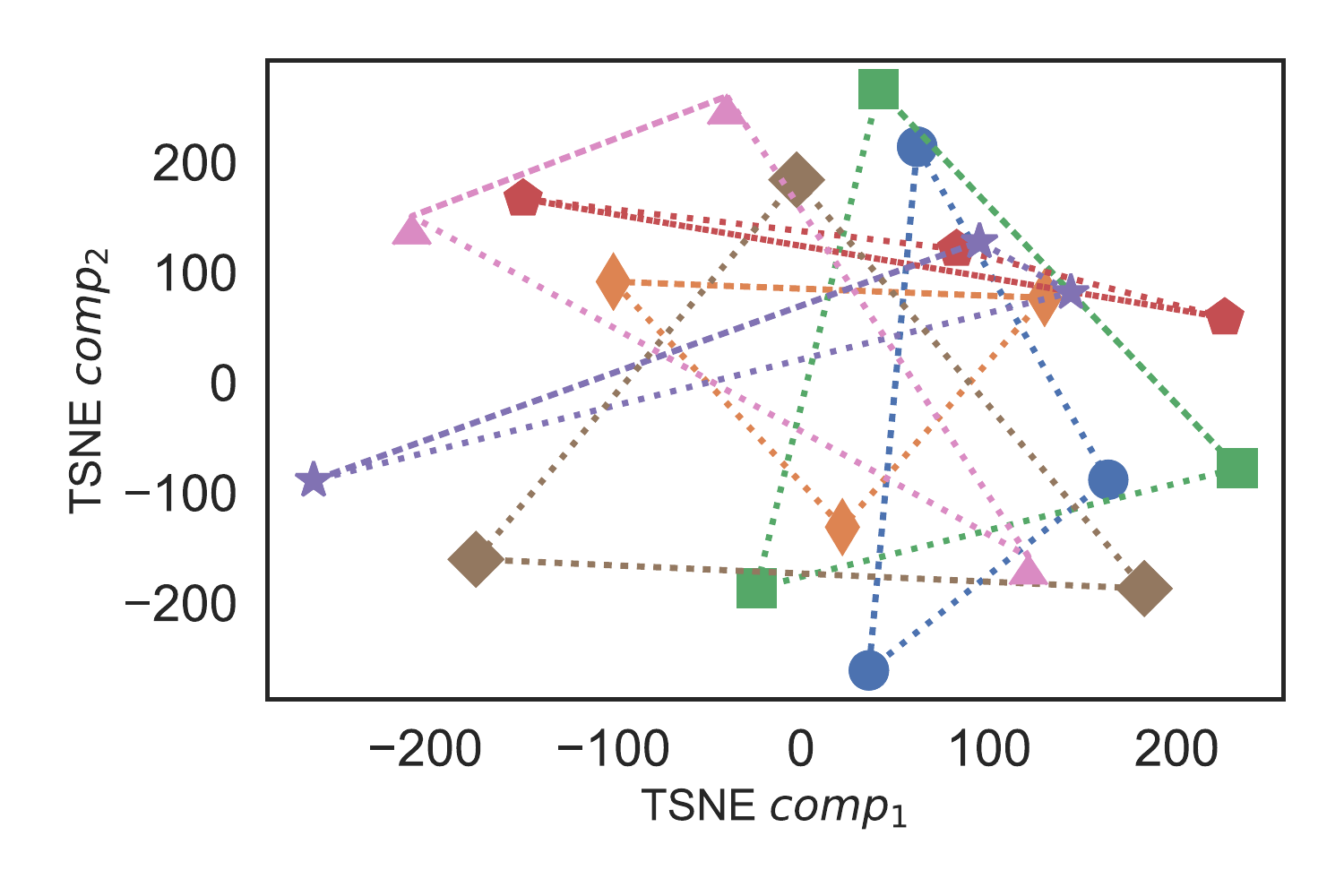}
    \caption{Task-oriented feature projection using t-SNE on NYU-v2 dataset. The triangle points having the same color refer to segmentation, depth, and surface normal features for the same input. The distant feature position for the same input verifies task privacy in t-SNE.}
    \label{fig:tsne}
    \end{minipage}\hfill
    \begin{minipage}{0.45\textwidth}
        \centering
        \includegraphics[width=0.8\linewidth]{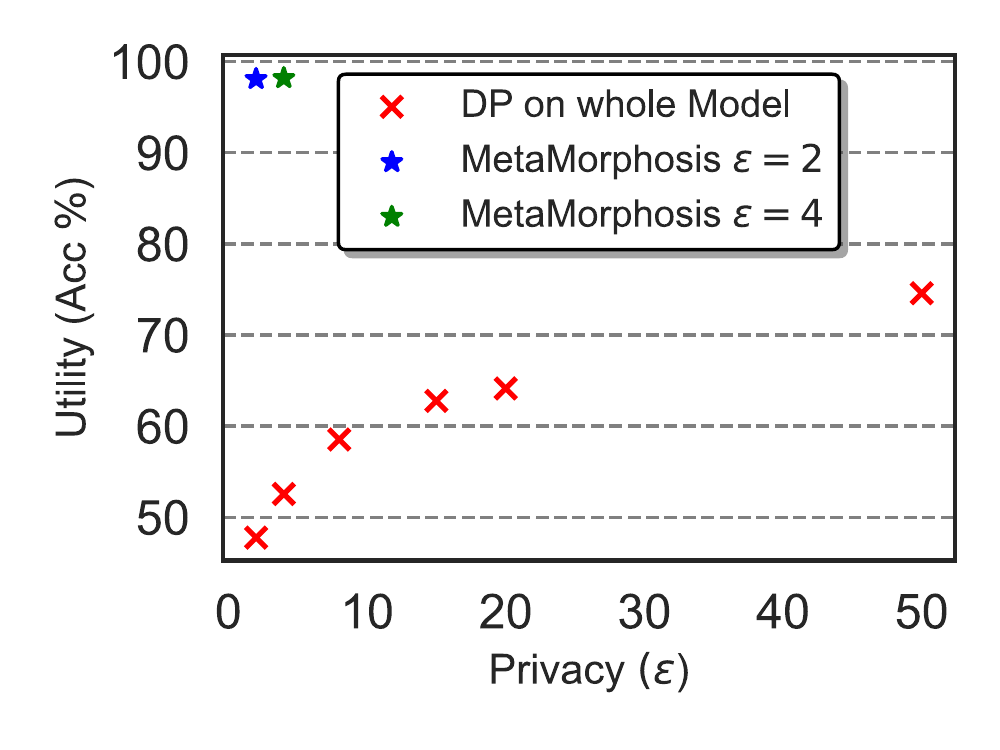}
    \caption{\my achieves better utility whereas respecting input obfuscation. Higher privacy ensures if $\epsilon \to 0$. So, the best points having high utility and high privacy locates in the top left quadrant.}
    \label{fig:eps_acc}
    \end{minipage}
\end{figure}
\begin{table*}[!t]
\caption{Test results on NYU-v2 dataset. In spite of imposing task privacy, \my achieves almost the same performance as~\cite{li2020knowledge}. STL refers to the results of single-task learning models.}
\label{tab:nyuv2}
\centering
\resizebox{\textwidth}{!} {%
\begin{tabular}{ccccc|cc|ccccc}
\toprule
\multirow{2}{*}{Model} & Size & \multirow{2}{*}{Methods} & \multicolumn{2}{c|}{Segmentation} & \multicolumn{2}{c|}{Depth} & \multicolumn{5}{c} {Surface Normal}\\
\cline{4-12}
 &(MB) &  & mIoU $(\uparrow)$ & Pix Acc $(\uparrow)$ & Abs Err $(\downarrow)$ & Rel Err $(\downarrow)$ & Mean $(\downarrow)$ & Median $(\downarrow)$ & 11.25 $(\uparrow)$  & 22.5 $(\uparrow)$ & 30 $(\uparrow)$ \\
 
\midrule

\multirow{3}{*}{SegNet}   & 300.90  & STL & 17.32 & 55.70 & 0.6577 & 0.2828 & 29.99 & 23.81 & 24.31 & 48.06 & 60.05\\
\cline{2-12}
   & 300.90  & KD-MTL~\cite{li2020knowledge}  & 18.75 & 58.02 & 0.5780 & 0.2467 & 29.40 & 23.71 & 24.33 & 48.22 & 60.45\\
    
  & 310.8 & \my  & 18.14 & 57.03 & 0.5867 & 0.2498 & 30.47 & 24.73 & 22.92 & 46.50 & 58.62\\
  

\bottomrule
\end{tabular}
}%
\end{table*}
\begin{table*}[!t]
\caption{NYU-v2~\cite{SilbermanECCV12} task-privacy evaluation by interchanging metamorphosis module.}
\label{tab:nyuv2-tp}
\centering
\resizebox{\textwidth}{!}{
\begin{tabular}{cccc|cc|ccccc}
\toprule
\my & Methods & \multicolumn{2}{c|}{Segmentation} & \multicolumn{2}{c|}{Depth} & \multicolumn{5}{c} {Surface Normal}\\
 Module &  & mIoU $(\uparrow)$ & Pix Acc $(\uparrow)$ & Abs Err $(\downarrow)$ & Rel Err $(\downarrow)$ & Mean $(\downarrow)$ & Median $(\downarrow)$ & 11.25 $(\uparrow)$  & 22.5 $(\uparrow)$ & 30 $(\uparrow)$  \\
\midrule
--- & \my  & 18.14 & 57.03 & 0.5867 & 0.2498 & 30.47 & 24.73 & 22.92 & 46.50 & 58.62\\
\midrule
    Segmentation  & \my  & 18.14 & 57.03 & \underline{1.2541} & \underline{0.5014} & \underline{51.97} & \underline{51.37} & \underline{1.74} & \underline{9.14} & \underline{17.89}\\
    
   Depth  &   &  \underline{4.27} & \underline{18.28} & 0.5867 & 0.2498 & \underline{54.77} & \underline{54.38} & \underline{4.41} & \underline{14.41} & \underline{22.02}\\

   Surface-normal  &  & \underline{3.37} & \underline{16.56} & \underline{1.8694} & \underline{0.7843} & 30.47 & 24.73 & 22.92 & 46.50 & 58.62\\
\midrule

\end{tabular}
}%
\end{table*}
We also observe the qualitative results of CityScapes using task privacy as shown in Figure~\ref{fig:quant-cs}. The segmentation and depth estimation outputs are almost obscured if respective features are not used for respective tasks. 
To evaluate more complicated tasks, we consider adding surface normal estimation with the segmentation and depth tasks and impose task privacy. We use NYU-v2 dataset in this regard. We have found similar results as on Cityscapes dataset. In NYU-v2, we also observe similar performance as compared to KD-MTL~\cite{li2020knowledge} with a little deflection in performance metric ($\sim<1\%$ for segmentation and depth estimation, and $\sim<2\%$ for surface normal estimation as shown in Table~\ref{tab:nyuv2}. We also evaluate task privacy on NYU-v2 by interchanging the metamorphosis modules as shown in Table~\ref{tab:nyuv2-tp}. The mIoU for segmentation drops down to $3.37\sim4.27$, the absolute depth error rises to >1, and the mean value of surface normal goes high from 30.47 to $51.97\sim54.77$.To evaluate the task-specific feature projection, we investigate the inference of the model trained on NYU-v2 and project the three task features using t-SNE representation as shown in Figure~\ref{fig:tsne}. The task features for each input are projected by training them using t-SNE. We show the task projection points in the same color and form a triangle to observe how separate they are. The higher area of the triangle means a higher distance. The component values of t-SNE additionally illustrate the distance among feature projections for each input. 

\begin{figure*}
    \centering
    \includegraphics[width=0.9\textwidth]{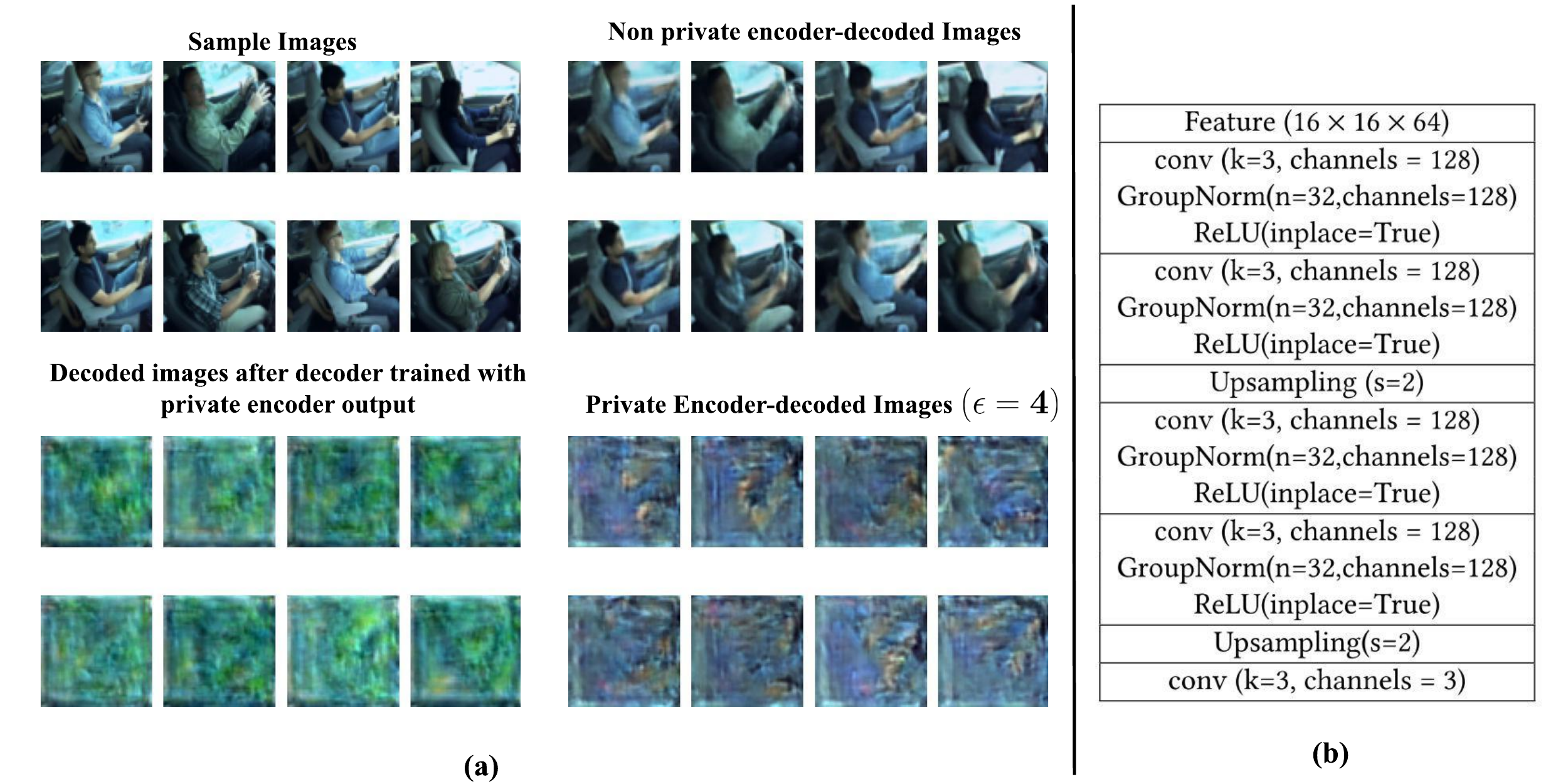}
    \caption{Input obfuscation on the StateFarm dataset. (a) Top left: Sample images. Top right: images decoded from a non-private encoder. Bottom right: images decoded from a private encoder trained with differential privacy having $\epsilon = 4, \delta = 1e^{-4}$ and a decoder (Figure~\ref{fig:dp-effective}b). Bottom left: Images generated by a decoder after training using the same private encoder. (b) Decoder model by the attacker for image reconstruction attack on private features.}
    \label{fig:dp-effective}
\end{figure*}

\textbf{StateFarm:} To evaluate \my in achieving task utility and input obfuscation, we use the distracted driver recognition task having 10 classes. At first, we impose differential privacy (DP) into the model encoder and classifier part. By varying the $\epsilon$, we compute the distracted behavior recognition accuracy. In Figure~\ref{fig:eps_acc}, we observe the accuracy drops with the increase in privacy (In DP, the $\epsilon \to 0$ ensured higher privacy and vice-versa). According to Algorithm~\ref{algo-meta}, to ensure input obfuscation, by adding DP-guarantee to the encoder side only, \my achieves both utility and privacy.
We also evaluate differential privacy qualitatively. In Figure~\ref{fig:dp-effective}, reconstruction of the Statefarm dataset is performed using an encoder and decoder. Using the encoder features, the distracted driver recognition task is performed. The decoder succeeds in decoding the image. Then, we train the encoder using differential privacy. In this case, the decoder fails to reconstruct images. Even using the private encoder, we train a decoder to reconstruct the image. Even after training, the decoder failed, but we got 98.69\% accuracy for the intended distracted driver recognition task. In Figure~\ref{fig:dp-effective}, we observe the DP-guarantee of obfuscating deep features in spite of training a decoder with the obfuscated features.

In addition to a private attribute, we first train the encoder and the private attribute task jointly using DP on the encoder. After the models are trained, then we train the intended classification task using MetaMorphosis, where a task-transformer module is trained using DP, and it generates distinct features for the distracted driver recognition task from private features by adding MS-SSIM to the loss function. In this way, the full process will maintain content-task privacy. Table~\ref{tab:statefarm} shows the evaluation of the final output of driver identity recognition and distracted driver behavior recognition tasks. After training with MetaMorphosis, for driver identity recognition, we got 100\%, and for distracted driver behavior detection, we got 98.69\% accuracy. Then by interchanging the task-transformer module, we compute the accuracy of each task which implies task privacy. We observe only 1.49\% accuracy if an intruder use behavior features for driver identity classification. In comparison to deepObfuscator~\cite{li2021deepobfuscator}, considering behavior features as general features to shared (as private features will be hidden in feature producer), \my achieves only 1.49\% accuracy for driver identity recognition whereas, for deepObfuscator, it achieves 30.38\% accuracy. So, \my ensures more privacy in obfuscating private attributes.

\begin{table}[!t]
\caption{Test results on input obfuscation and task-privacy on Statefarm dataset}
\label{tab:statefarm}
\centering
\resizebox{\columnwidth}{!}{
\begin{tabular}{cccc}
\toprule
Task Metamorphosis & Classifier &DeepObfuscator~\cite{li2021deepobfuscator} & \my \\
 Module &  & &  \\
\midrule
Identity & Identity & 99.97 & 100.00 \\
Behavior & Identity & 30.28 & \textbf{1.49}\\
Behavior & Behavior & 98.32 & 98.69\\
Identity & Behavior & --- & \textbf{10.34}\\
\bottomrule
\end{tabular}}
\end{table}

\begin{table}[!htbp]

\caption{Test results on task privacy and input obfuscation on CelebA}
\label{tab:gen-smile}
\centering
\resizebox{\columnwidth}{!}{
\begin{tabular}{ccccc}
\toprule
split point & Task Metamorphosis & Classifier &DeepObfuscator & \my \\
 & Module &  & &  \\
\midrule
5 &Gender & Gender & --- & 95.94 \\
&Smile & Gender & 55.85 & \textbf{34.56}\\
&Smile & Smile & 89.52 & 89.89\\
&Gender & Smile & --- & \textbf{42.49}\\
\bottomrule
\end{tabular}
}
\end{table}

\textbf{CelebA:} To validate the task privacy and input obfuscation jointly, We have experimented with CelebA dataset. We consider two scenarios (1) smile, gender, and (2) smile, gender, cheekbone classification where gender is a private attribute and input obfuscation is imposed. To achieve this, according to Algorithm~\ref{algo-meta}, we joint train the encoder and the private gender classifier first. For this, we also use a comparatively smaller model, ResNet-18, and split it into different positions to build the encoder and the classifier. Without loss of generality, we use the same classifier model size for all tasks. After training of encoder with DP and the private classifier, we use the task-privacy loss to train the classifier for smile for case (1) and the smile and cheekbone classifier jointly for case (2). Then, we test the performance of all tasks and task privacy by interchanging the task-metamorphosis modules. It is to be noted that gender features are created so that we can make other attributes' features distinct from the private features, and this private attribute feature will be hidden and kept on the producer side. As in MetaMorphosis, we show that one private attribute defined for one task may be defined as non-private by another task. In Table~\ref{tab:gen-smile}, we consider gender as private and the smile classification as the intended classification task. Both private and non-private classifiers achieve almost similar performance as DeepObfuscator~\cite{li2021deepobfuscator} but hide privacy information better (21.29\% less accuracy than DeepObfuscator). In this case, \my achieved 34.56\% accuracy while doing gender classification using the smile classifier. The reason for a bit increase in gender accuracy with one task (smile) and two tasks (smile and cheekbone) indicates an intrinsic correlation between the two tasks as discussed by recent literature~\cite{li2021deepobfuscator}. In this case, \my diminishes the correlation more than DeepObfuscator. 

        
        
\begin{figure}[!htpb]
    \centering
    \includegraphics[width=0.9\columnwidth]{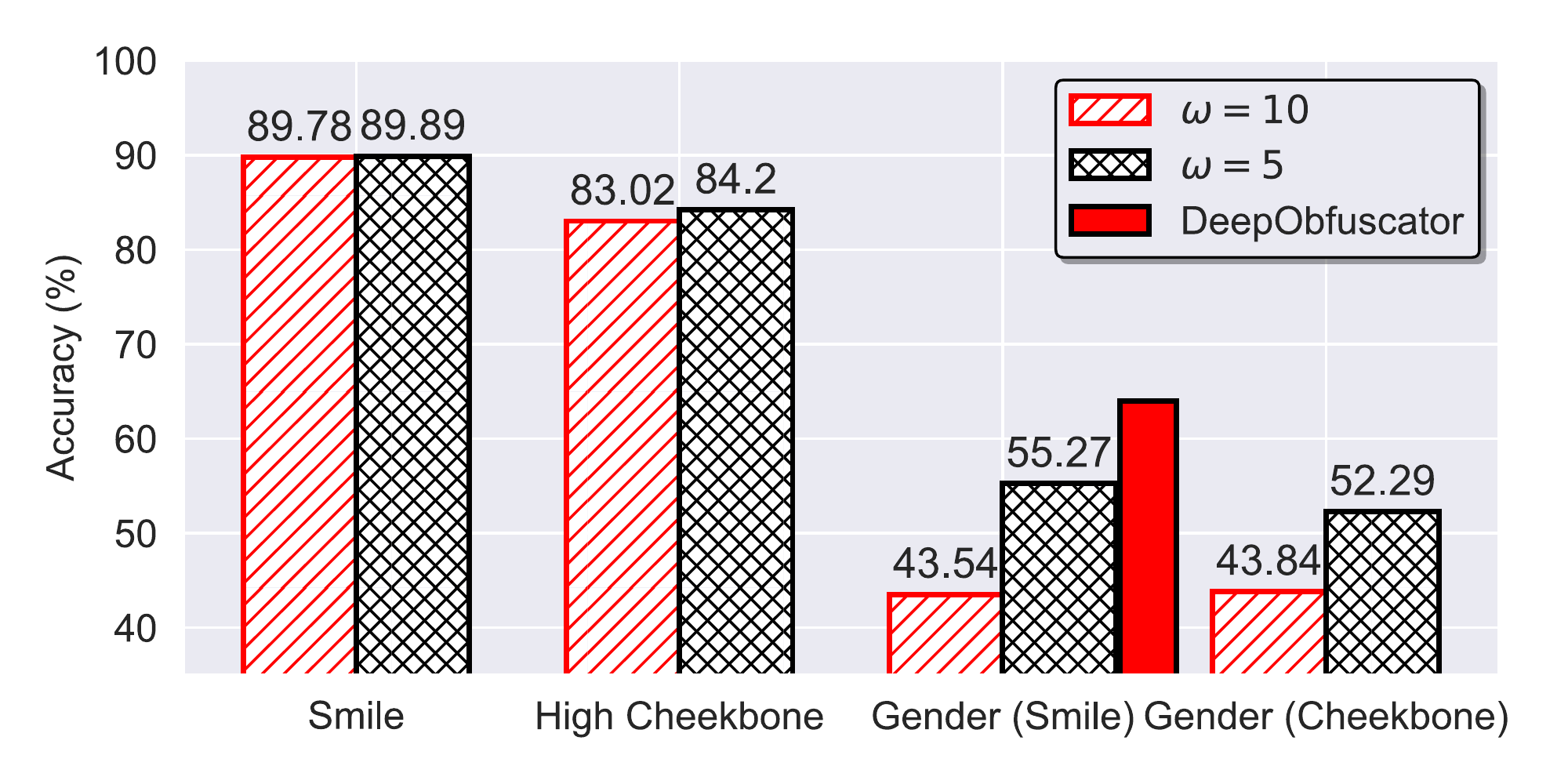}
    \caption{Smile and cheekbone classification from CelebA while obfuscating gender attribute using MetaMorphosis. }
    \label{fig:smile-cheek}
\end{figure}
We investigate the second scenario where the number of intended tasks is two. In this, we adopt smile and cheekbone classification as two intended tasks. In this scenario, the privacy requirements are similar to the previous one i.e. input obfuscation and task privacy. \my classifies smile and cheekbone while obfuscating the gender attribute and input. In Figure~\ref{fig:smile-cheek}, we achieve $89.78\sim89.89\%$ accuracy for smile and $83.02\sim84.2\%$ accuracy for cheekbone classification using a variety of weight $\omega$ of task-privacy while ensuring input obfuscation using DP. \my has achieved similar results for intended task classification but hides privacy $8.69\%\sim20.42\%$ more than DeepObfuscator~\cite{li2021deepobfuscator} using smile classification task features and $11.69\%\sim 20.12\%$ using cheekbone classification task features. Relation between $\omega$ and accuracy basically depends on task-correlation and is interesting to investigate which we keep as our future work.

\section{System Deployment}\label{sec:system}
\begin{figure}
    \centering
    \includegraphics[width=0.9\columnwidth]{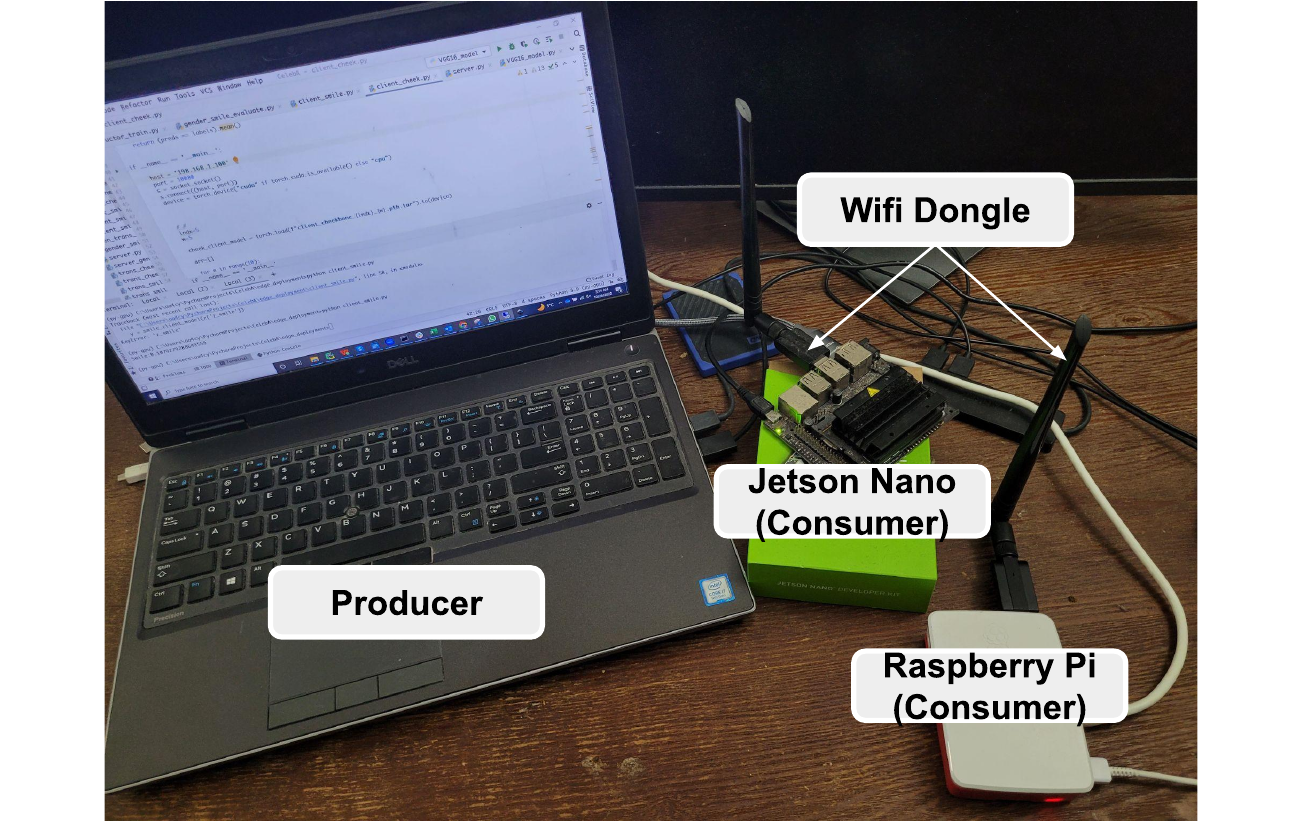}
    \caption{Producer-Consumer Deployment}
    \label{fig:system}
\end{figure}

For system deployment, the producer will send features, and the consumer will produce the outputs. But to build such a setup, the consumer needs to know the ground truth to compute the loss function. If the features and labels are shared for joint training, the attacker can eavesdrop on the features and design any system to capture the feature output relationship. Instead, the labels are sent at a specific time using an encryption key. Then, the features can be shared to do the training task. To replicate the scenario, we consider a Quadro RTX 4000 as the producer (server) and a Raspberry-pi as well as a Jetson Nano as the consumers (client) in Figure~\ref{fig:system}. At the forward pass of training, the producer sends the intermediate features to the consumers. The consumers produce the output, compute the loss function and send the computed gradient in the backward pass. Based on the gradient, the producer updates its parameters. Based on input obfuscation, the producer is trained with a DP-optimizer. 

We demonstrate a split neural network using ResNet-18 model with different indices as split points and execute the training. As an MLaaS platform, the service provider should offer features such that the consumer can do the task with little effort. With the smaller consumer model size, the round trip time will be lower for the accumulation of gradients by the producer. Table~\ref{tab:rtt} shows the higher the round trip time with the higher feature, the larger the client model size. Due to the usage of GPU by Nano, the difference between the round-trip time of Jetson Nano and Raspberry-Pi is significant. 

\begin{table}[!t]
\centering
\caption{Round trip time of sending features and collecting the gradients vs the consumer (client) model size vs the intermediate feature size.}
\label{tab:rtt}
\resizebox{\columnwidth}{!}{%
\begin{tabular}{cccccc}
\toprule
Server Model & Client Model & Raspberry Pi   & Jetson nano & Feature  & Feature size \\
(MB)         & (MB)         & RTT (ms)   &  RTT (ms)   &     & KB           \\
\midrule
42.66        & 0.02         & 3.95   & 0.73  &512 $\times$ 1 $\times$ 1      & 1.99         \\
10.64        & 32.04        & 112.67 & 10.11 & 256 $\times$ 4 $\times$ 4  & 14.86        \\
0.61         & 42.07        & 173.23 & 18.60 & 64 $\times$ 16 $\times$ 16 & 65.92      \\
\bottomrule
\end{tabular}%
}
\end{table}
We also investigate the effect of adding a task metamrophosis module to the overall server latency. The latency of the metamorphosis module depends on the input shared by the encoder and its size. Figure~\ref{tab:latency-effect} shows the effect of inference latency of adding the metamorphosis module with the encoder for running the smile classification task with the encoder, smile metamorphosis module, and the classifier. The little difference in latency proves the metamorphosis module to be a lightweight one.
\begin{table}[!t]
    \centering
    \caption{Effect of \my module to Server latency}
    \label{tab:latency-effect}
    \resizebox{0.9\columnwidth}{!}{
    \begin{tabular}{ccccccc}
    \toprule
     MetaMorphosis &  Model & Split & Server & \my  & Server (ms) \\
     Module &  & Index & Size & Module Size & Latency (ms) \\
        \midrule
     \xmark &   ResNet-18 & 5 & 640.60 KB & 54.5 KB & 0.068 \\
     \cmark &    &  & & &   0.106 \\
     \midrule
     \xmark &   ResNet-18 & 7  & 11.20 MB &791.80 KB& 0.244 \\
     \cmark &    &  & & &  0.265 \\
     \midrule
     \xmark &   ResNet-18 & 8  & 44.70 MB & 3.20 MB & 0.293 \\
     \cmark &    &  & & &   0.359 \\
         \bottomrule
    \end{tabular}}
    
\end{table}

\section{Ablation Study}\label{sec:ablation}
We evaluate the performance of the Cross-SEC module, and we also experiment without crossing the connections after getting the attention of one portion of the features. In Table~\ref{tab:imp_cross_sec}, joint training of segmentation and depth estimation is done where the task-privacy module is the Cross-SEC module and SEC module where no cross-connection between features occurs. Cross-SEC morphosis module performs better in achieving the metrics for segmentation and depth estimation than the SEC module. Regarding task privacy, cross-SEC achieves lower pixel accuracy, mIOU for segmentation with depth features, and lower absolute error for depth estimation with segmentation features than SEC module.

\begin{table}[!htbp]
\caption{Importance of Cross-SEC module over SEC module without cross attention}
\label{tab:imp_cross_sec}
\centering
\resizebox{0.9\columnwidth}{!}{
\begin{tabular}{cccc|cc}
\toprule
\my & Methods & \multicolumn{2}{c|}{Segmentation} & \multicolumn{2}{c}{Depth} \\
 Module &  & mIoU $(\uparrow)$ & Pix Acc $(\uparrow)$ & Abs Err $(\downarrow)$ & Rel Err $(\downarrow)$ \\
\midrule
 --- &  MTL-SEC & 57.79 & 93.39 & 0.0148 & 45.07\\
\midrule
Segmentation & MTL-SEC & 57.79 & 93.39 & \underline{0.1022} & \underline{110.09}\\
Depth &  & \underline{3.92} & \underline{23.97} & 0.0148 & 45.07\\
\midrule
 --- &  MTL-Cross-SEC & \textbf{59.79} & \textbf{93.49} & \textbf{0.0141} & \textbf{31.89}\\
\midrule
Segmentation & MTL-Cross-SEC & 59.79 & 93.49 & \underline{0.1079} & \underline{99.07}\\
Depth &  & \underline{1.47} & \underline{7.33} & 0.0141 & 31.89\\
\bottomrule

\end{tabular}
}%
\end{table}

\begin{table}[!htbp]
\caption{Input-attribute obfuscation trade-off}
\label{tab:input-attr}
\centering
\resizebox{0.9\columnwidth}{!}{
\begin{tabular}{ccccccc}
\toprule
Method & Model & Provider  & Classifier  & \my & Classifier & Accuracy (\%)\\
&  & Size & Size   & Module &  & \\
&  & (MB) &  (MB)  &  &  & \\
\midrule
DeepObfuscator~\cite{li2021deepobfuscator} & VGG16 & 1.0 & 536 & universal & gender & 55.85\\
\midrule
\my & ResNet18 & 3.00 & 42.00 & gender & gender & 95.44\\
 &  &  &  & smile & smile & 87.78\\
  &  &  &  & gender & smile & 42.69\\
 &  &  &  & smile & gender & 45.14\\
 \midrule
 &  & 11.97 & 33.60 & gender & gender & 94.19\\
 &  &  &  & smile & smile & 82.09\\
  &  &  &  & gender & smile & 42.56\\
 &  &  &  & smile & gender & 61.35\\
 \midrule
 &  & 47.9 & 0.006 & gender & gender & 92.74\\
 &  &  &  & smile & smile & 58.83\\
  &  &  &  & gender & smile & 42.33\\
 &  &  &  & smile & gender & 38.65\\
\bottomrule
\end{tabular}
}
\end{table}

To observe the trade-off between input and privacy attribute obfuscation, we change the encoder and classifier size by changing the split index of the ResNet-18 model. We identify an increase in privacy attribute leakage with the decrease of the classifier model size and expansion of the encoder model size (Table~\ref{tab:input-attr}). We have found the gender classifier accuracy 45.14\%, and 61.25 with lowering the classifier size from 42 MB to 33.6 MB. It is even worse for a classifier having 0.006 MB. We observe that with the increase of the encoder model and the decrease of the classifier model, it is difficult for the intended classification task to meet the input obfuscation and task privacy together. As more noise is fed to the encoder model to maintain the $\epsilon$-DP while training, less performance is desired with the decreased classifier model size, as also evident from Figure~\ref{fig:eps_acc}. DeepObfuscator used VGG-16 where only 1 MB portion is defined as feature provider, and the 536 MB is designated to the intended class classifier whereas in MetaMorphosis, even using a small model ResNet-18, with higher encoder size, we achieve almost the same accuracy as DeepObfuscator and hides privacy attribute better by lowering gender classification task. Finding the optimal split index between the encoder and the task classifier is an interesting area to achieve input obfuscation and task privacy. We have kept this discussion as our future work.
\section{Related Work}\label{sec:relatedwork}
Various methods for solving complex segmentation and depth estimation-based multi-task learning are discussed in the literature~\cite{bhat2021adabins,wang2019lednet,nimi2022chimera}. A knowledge distillation technique is proposed by Liu et al.~\cite{liu2019structured}  specifically for semantic segmentation tasks. Nguyen et al.~\cite{nguyen2019multi} proposed a convolutional neural network to identify modified images, and the trained network can give a segmented mask for the modified region. An empirical study has been conducted by Standley et al.~\cite{standley2020tasks} to identify the factors that influence the performance of multi-task learning and proposed a framework to limit the number of multi-task models based on the correlation of tasks.
SSIM~\cite{wang2004image} provides an image quality assessment metric called structural similarity (SSIM) to evaluate the similarity between two images. It can also be used as a loss function to impose dissimilarity between features by shifting the value close to zero.
Attention modules are proposed in the literature~\cite{zhang2021sa,wang2020eca} to capture important features for target accuracy without dimensionality reduction. Chen et al.~\cite{chen2018gradnorm} propose a gradient normalization algorithm for training multi-task models to balance the training processes of different tasks. The algorithm improves accuracy and decreases the over-fitting effects for various kinds of tasks and on different datasets. Transformer-based cross-task attention mechanism~\cite{lopes2022cross} projects the features of one task to another. But the notion of distinct feature generation to achieve privacy is absent. In collaborative intelligence, to build a more efficient system, many layer output compression methods~\cite{cohen2020lightweight, choi2018deep, yao2020deep} and gradient compression methods~\cite{vogels2019powersgd} are suggested. Other~\cite{zhang2021msfc, wang2021deep} focuses on multi-task feature compression. These compression techniques, referred to as Video Coding for Machine (VCM)~\cite{yang2021video}, aims to reduce the communication overhead while maintaining the system performance, while many efforts~\cite{iandola2016squeezenet, zhang2021sa, li2019selective, zhang2022resnest, pan2022integration, hu2018squeeze} are also devoted to optimizing the computational overhead. On the other hand, to build a good collaborative intelligent system, besides improving its efficiency, privacy-preserving feature encoding schemes also need to protect the privacy of data holders. In the case of differential privacy in deep learning, the DP-SGD algorithm was proposed by Abadi et al.~\cite{abadi2016deep}. Many variants of differential privacy, such as label differential privacy, are discussed in~\cite{ghazi2021deep}. Table~\ref{tab:lit-compare} illustrates the comparison of \my with recent similar literature.
%
%

\section{Conclusion}\label{sec:conclusion}
In this paper, we propose \my that enables input obfuscation and task privacy for multi-task learning in a collaborative intelligence setup. In this paper, the main focus lies in sharing data and computation securely between a deep feature provider-based MLaas platform and a number of consumers who subscribe to the provider according to interest. To ensure this, \my has gone through a two-phase training scheme where the first phase ensures input privacy and private attribute privacy. Then the second training phase ensures task privacy among shared tasks through a unique squeeze-excitation based \my module. Experimental results on different domain datasets show the supremacy of \my over recent multi-task and adversarial learning methods. The \my also positively affects the sequential addition of new tasks in a multi-task environment because of its two-phase training scheme. This paper opens up some questions and disadvantages of having a universal feature for a split learning system as well as in a split federated learning system. As the performance of a federated learning system mostly depends on the \emph{honesty} of the clients, the intuitive creation of task features despite the task \emph{correlation} is still challenging with the increase in the number of tasks. In the future, we will focus on how task-variant features can be used to enable more privacy in federated learning systems. 

\begin{acks}
We thank our anonymous reviewers and our shepherd Mimi Xie for their valuable feedback.  
\end{acks}


\bibliographystyle{ACM-Reference-Format}
\bibliography{ref}


\begin{thebibliography}{48}


\ifx \showCODEN    \undefined \def \showCODEN     #1{\unskip}     \fi
\ifx \showDOI      \undefined \def \showDOI       #1{#1}\fi
\ifx \showISBNx    \undefined \def \showISBNx     #1{\unskip}     \fi
\ifx \showISBNxiii \undefined \def \showISBNxiii  #1{\unskip}     \fi
\ifx \showISSN     \undefined \def \showISSN      #1{\unskip}     \fi
\ifx \showLCCN     \undefined \def \showLCCN      #1{\unskip}     \fi
\ifx \shownote     \undefined \def \shownote      #1{#1}          \fi
\ifx \showarticletitle \undefined \def \showarticletitle #1{#1}   \fi
\ifx \showURL      \undefined \def \showURL       {\relax}        \fi
\providecommand\bibfield[2]{#2}
\providecommand\bibinfo[2]{#2}
\providecommand\natexlab[1]{#1}
\providecommand\showeprint[2][]{arXiv:#2}

\bibitem[Abadi et~al\mbox{.}(2016)]%
        {abadi2016deep}
\bibfield{author}{\bibinfo{person}{Martin Abadi}, \bibinfo{person}{Andy Chu},
  \bibinfo{person}{Ian Goodfellow}, \bibinfo{person}{H~Brendan McMahan},
  \bibinfo{person}{Ilya Mironov}, \bibinfo{person}{Kunal Talwar}, {and}
  \bibinfo{person}{Li Zhang}.} \bibinfo{year}{2016}\natexlab{}.
\newblock \showarticletitle{Deep learning with differential privacy}. In
  \bibinfo{booktitle}{\emph{Proceedings of the 2016 ACM SIGSAC conference on
  computer and communications security}}. \bibinfo{pages}{308--318}.
\newblock


\bibitem[Arefeen et~al\mbox{.}(2021a)]%
        {arefeen2021transjury}
\bibfield{author}{\bibinfo{person}{Md~Adnan Arefeen},
  \bibinfo{person}{Sumaiya~Tabassum Nimi}, \bibinfo{person}{Md~Yusuf~Sarwar
  Uddin}, {and} \bibinfo{person}{Yugyung Lee}.}
  \bibinfo{year}{2021}\natexlab{a}.
\newblock \showarticletitle{TransJury: Towards Explainable Transfer Learning
  through Selection of Layers from Deep Neural Networks}. In
  \bibinfo{booktitle}{\emph{2021 IEEE International Conference on Big Data (Big
  Data)}}. IEEE, \bibinfo{pages}{978--984}.
\newblock


\bibitem[Arefeen et~al\mbox{.}(2021b)]%
        {arefeen2021lightweight}
\bibfield{author}{\bibinfo{person}{Md~Adnan Arefeen},
  \bibinfo{person}{Sumaiya~Tabassum Nimi}, \bibinfo{person}{Md~Yusuf~Sarwar
  Uddin}, {and} \bibinfo{person}{Zhu Li}.} \bibinfo{year}{2021}\natexlab{b}.
\newblock \showarticletitle{A lightweight relu-based feature fusion for aerial
  scene classification}. In \bibinfo{booktitle}{\emph{2021 IEEE International
  Conference on Image Processing (ICIP)}}. IEEE, \bibinfo{pages}{3857--3861}.
\newblock


\bibitem[Bhat et~al\mbox{.}(2021)]%
        {bhat2021adabins}
\bibfield{author}{\bibinfo{person}{Shariq~Farooq Bhat},
  \bibinfo{person}{Ibraheem Alhashim}, {and} \bibinfo{person}{Peter Wonka}.}
  \bibinfo{year}{2021}\natexlab{}.
\newblock \showarticletitle{Adabins: Depth estimation using adaptive bins}. In
  \bibinfo{booktitle}{\emph{Proceedings of the IEEE/CVF Conference on Computer
  Vision and Pattern Recognition}}. \bibinfo{pages}{4009--4018}.
\newblock


\bibitem[Chen et~al\mbox{.}(2018b)]%
        {chen2018encoder}
\bibfield{author}{\bibinfo{person}{Liang-Chieh Chen}, \bibinfo{person}{Yukun
  Zhu}, \bibinfo{person}{George Papandreou}, \bibinfo{person}{Florian Schroff},
  {and} \bibinfo{person}{Hartwig Adam}.} \bibinfo{year}{2018}\natexlab{b}.
\newblock \showarticletitle{Encoder-decoder with atrous separable convolution
  for semantic image segmentation}. In \bibinfo{booktitle}{\emph{Proceedings of
  the European conference on computer vision (ECCV)}}.
  \bibinfo{pages}{801--818}.
\newblock


\bibitem[Chen et~al\mbox{.}(2018a)]%
        {chen2018gradnorm}
\bibfield{author}{\bibinfo{person}{Zhao Chen}, \bibinfo{person}{Vijay
  Badrinarayanan}, \bibinfo{person}{Chen-Yu Lee}, {and} \bibinfo{person}{Andrew
  Rabinovich}.} \bibinfo{year}{2018}\natexlab{a}.
\newblock \showarticletitle{Gradnorm: Gradient normalization for adaptive loss
  balancing in deep multitask networks}. In
  \bibinfo{booktitle}{\emph{International conference on machine learning}}.
  PMLR, \bibinfo{pages}{794--803}.
\newblock


\bibitem[Choi and Baji{\'c}(2018)]%
        {choi2018deep}
\bibfield{author}{\bibinfo{person}{Hyomin Choi} {and} \bibinfo{person}{Ivan~V
  Baji{\'c}}.} \bibinfo{year}{2018}\natexlab{}.
\newblock \showarticletitle{Deep feature compression for collaborative object
  detection}. In \bibinfo{booktitle}{\emph{2018 25th IEEE International
  Conference on Image Processing (ICIP)}}. IEEE, \bibinfo{pages}{3743--3747}.
\newblock


\bibitem[Cohen et~al\mbox{.}(2020)]%
        {cohen2020lightweight}
\bibfield{author}{\bibinfo{person}{Robert~A Cohen}, \bibinfo{person}{Hyomin
  Choi}, {and} \bibinfo{person}{Ivan~V Baji{\'c}}.}
  \bibinfo{year}{2020}\natexlab{}.
\newblock \showarticletitle{Lightweight compression of neural network feature
  tensors for collaborative intelligence}. In \bibinfo{booktitle}{\emph{2020
  IEEE International Conference on Multimedia and Expo (ICME)}}. IEEE,
  \bibinfo{pages}{1--6}.
\newblock


\bibitem[Cordts et~al\mbox{.}(2016)]%
        {cordts2016cityscapes}
\bibfield{author}{\bibinfo{person}{Marius Cordts}, \bibinfo{person}{Mohamed
  Omran}, \bibinfo{person}{Sebastian Ramos}, \bibinfo{person}{Timo Rehfeld},
  \bibinfo{person}{Markus Enzweiler}, \bibinfo{person}{Rodrigo Benenson},
  \bibinfo{person}{Uwe Franke}, \bibinfo{person}{Stefan Roth}, {and}
  \bibinfo{person}{Bernt Schiele}.} \bibinfo{year}{2016}\natexlab{}.
\newblock \showarticletitle{The cityscapes dataset for semantic urban scene
  understanding}. In \bibinfo{booktitle}{\emph{Proceedings of the IEEE
  conference on computer vision and pattern recognition}}.
  \bibinfo{pages}{3213--3223}.
\newblock


\bibitem[Ding et~al\mbox{.}(2020)]%
        {ding2020privacy}
\bibfield{author}{\bibinfo{person}{Xiaofeng Ding}, \bibinfo{person}{Hongbiao
  Fang}, \bibinfo{person}{Zhilin Zhang}, \bibinfo{person}{Kim-Kwang~Raymond
  Choo}, {and} \bibinfo{person}{Hai Jin}.} \bibinfo{year}{2020}\natexlab{}.
\newblock \showarticletitle{Privacy-preserving feature extraction via
  adversarial training}.
\newblock \bibinfo{journal}{\emph{IEEE Transactions on Knowledge and Data
  Engineering}} (\bibinfo{year}{2020}).
\newblock


\bibitem[Elhassan et~al\mbox{.}(2021)]%
        {elhassan2021dsanet}
\bibfield{author}{\bibinfo{person}{Mohammed~AM Elhassan},
  \bibinfo{person}{Chenxi Huang}, \bibinfo{person}{Chenhui Yang}, {and}
  \bibinfo{person}{Tewodros~Legesse Munea}.} \bibinfo{year}{2021}\natexlab{}.
\newblock \showarticletitle{DSANet: Dilated spatial attention for real-time
  semantic segmentation in urban street scenes}.
\newblock \bibinfo{journal}{\emph{Expert Systems with Applications}}
  \bibinfo{volume}{183} (\bibinfo{year}{2021}), \bibinfo{pages}{115090}.
\newblock


\bibitem[Ghazi et~al\mbox{.}(2021)]%
        {ghazi2021deep}
\bibfield{author}{\bibinfo{person}{Badih Ghazi}, \bibinfo{person}{Noah
  Golowich}, \bibinfo{person}{Ravi Kumar}, \bibinfo{person}{Pasin Manurangsi},
  {and} \bibinfo{person}{Chiyuan Zhang}.} \bibinfo{year}{2021}\natexlab{}.
\newblock \showarticletitle{Deep learning with label differential privacy}.
\newblock \bibinfo{journal}{\emph{Advances in Neural Information Processing
  Systems}}  \bibinfo{volume}{34} (\bibinfo{year}{2021}),
  \bibinfo{pages}{27131--27145}.
\newblock


\bibitem[{Google}(2022)]%
        {google}
\bibfield{author}{\bibinfo{person}{{Google}}.} \bibinfo{year}{2022}\natexlab{}.
\newblock \bibinfo{title}{Google video intelligence API.}
\newblock
\newblock
\newblock
\shownote{\url{https://cloud.google.com/video-intelligence/}}.


\bibitem[Hu et~al\mbox{.}(2018)]%
        {hu2018squeeze}
\bibfield{author}{\bibinfo{person}{Jie Hu}, \bibinfo{person}{Li Shen}, {and}
  \bibinfo{person}{Gang Sun}.} \bibinfo{year}{2018}\natexlab{}.
\newblock \showarticletitle{Squeeze-and-excitation networks}. In
  \bibinfo{booktitle}{\emph{Proceedings of the IEEE conference on computer
  vision and pattern recognition}}. \bibinfo{pages}{7132--7141}.
\newblock


\bibitem[Iandola et~al\mbox{.}(2016)]%
        {iandola2016squeezenet}
\bibfield{author}{\bibinfo{person}{Forrest~N Iandola}, \bibinfo{person}{Song
  Han}, \bibinfo{person}{Matthew~W Moskewicz}, \bibinfo{person}{Khalid Ashraf},
  \bibinfo{person}{William~J Dally}, {and} \bibinfo{person}{Kurt Keutzer}.}
  \bibinfo{year}{2016}\natexlab{}.
\newblock \showarticletitle{SqueezeNet: AlexNet-level accuracy with 50x fewer
  parameters and< 0.5 MB model size}.
\newblock \bibinfo{journal}{\emph{arXiv preprint arXiv:1602.07360}}
  (\bibinfo{year}{2016}).
\newblock


\bibitem[Khattar et~al\mbox{.}(2021)]%
        {khattar2021cross}
\bibfield{author}{\bibinfo{person}{Apoorv Khattar}, \bibinfo{person}{Srinidhi
  Hegde}, {and} \bibinfo{person}{Ramya Hebbalaguppe}.}
  \bibinfo{year}{2021}\natexlab{}.
\newblock \showarticletitle{Cross-domain multi-task learning for object
  detection and saliency estimation}. In \bibinfo{booktitle}{\emph{Proceedings
  of the IEEE/CVF Conference on Computer Vision and Pattern Recognition}}.
  \bibinfo{pages}{3639--3648}.
\newblock


\bibitem[Li et~al\mbox{.}(2020)]%
        {li2020tiprdc}
\bibfield{author}{\bibinfo{person}{Ang Li}, \bibinfo{person}{Yixiao Duan},
  \bibinfo{person}{Huanrui Yang}, \bibinfo{person}{Yiran Chen}, {and}
  \bibinfo{person}{Jianlei Yang}.} \bibinfo{year}{2020}\natexlab{}.
\newblock \showarticletitle{TIPRDC: task-independent privacy-respecting data
  crowdsourcing framework for deep learning with anonymized intermediate
  representations}. In \bibinfo{booktitle}{\emph{Proceedings of the 26th ACM
  SIGKDD International Conference on Knowledge Discovery \& Data Mining}}.
  \bibinfo{pages}{824--832}.
\newblock


\bibitem[Li et~al\mbox{.}(2021a)]%
        {li2021deepobfuscator}
\bibfield{author}{\bibinfo{person}{Ang Li}, \bibinfo{person}{Jiayi Guo},
  \bibinfo{person}{Huanrui Yang}, \bibinfo{person}{Flora~D Salim}, {and}
  \bibinfo{person}{Yiran Chen}.} \bibinfo{year}{2021}\natexlab{a}.
\newblock \showarticletitle{DeepObfuscator: Obfuscating intermediate
  representations with privacy-preserving adversarial learning on smartphones}.
  In \bibinfo{booktitle}{\emph{Proceedings of the International Conference on
  Internet-of-Things Design and Implementation}}. \bibinfo{pages}{28--39}.
\newblock


\bibitem[Li and Bilen(2020)]%
        {li2020knowledge}
\bibfield{author}{\bibinfo{person}{Wei-Hong Li} {and} \bibinfo{person}{Hakan
  Bilen}.} \bibinfo{year}{2020}\natexlab{}.
\newblock \showarticletitle{Knowledge distillation for multi-task learning}. In
  \bibinfo{booktitle}{\emph{European Conference on Computer Vision}}. Springer,
  \bibinfo{pages}{163--176}.
\newblock


\bibitem[Li et~al\mbox{.}(2021b)]%
        {li2021universal}
\bibfield{author}{\bibinfo{person}{Wei-Hong Li}, \bibinfo{person}{Xialei Liu},
  {and} \bibinfo{person}{Hakan Bilen}.} \bibinfo{year}{2021}\natexlab{b}.
\newblock \showarticletitle{Universal representation learning from multiple
  domains for few-shot classification}. In
  \bibinfo{booktitle}{\emph{Proceedings of the IEEE/CVF International
  Conference on Computer Vision}}. \bibinfo{pages}{9526--9535}.
\newblock


\bibitem[Li et~al\mbox{.}(2022)]%
        {li2022universal}
\bibfield{author}{\bibinfo{person}{Wei-Hong Li}, \bibinfo{person}{Xialei Liu},
  {and} \bibinfo{person}{Hakan Bilen}.} \bibinfo{year}{2022}\natexlab{}.
\newblock \showarticletitle{Universal Representations: A Unified Look at
  Multiple Task and Domain Learning}.
\newblock \bibinfo{journal}{\emph{arXiv preprint arXiv:2204.02744}}
  (\bibinfo{year}{2022}).
\newblock


\bibitem[Li et~al\mbox{.}(2019)]%
        {li2019selective}
\bibfield{author}{\bibinfo{person}{Xiang Li}, \bibinfo{person}{Wenhai Wang},
  \bibinfo{person}{Xiaolin Hu}, {and} \bibinfo{person}{Jian Yang}.}
  \bibinfo{year}{2019}\natexlab{}.
\newblock \showarticletitle{Selective kernel networks}. In
  \bibinfo{booktitle}{\emph{Proceedings of the IEEE/CVF conference on computer
  vision and pattern recognition}}. \bibinfo{pages}{510--519}.
\newblock


\bibitem[Liu et~al\mbox{.}(2019)]%
        {liu2019structured}
\bibfield{author}{\bibinfo{person}{Yifan Liu}, \bibinfo{person}{Ke Chen},
  \bibinfo{person}{Chris Liu}, \bibinfo{person}{Zengchang Qin},
  \bibinfo{person}{Zhenbo Luo}, {and} \bibinfo{person}{Jingdong Wang}.}
  \bibinfo{year}{2019}\natexlab{}.
\newblock \showarticletitle{Structured knowledge distillation for semantic
  segmentation}. In \bibinfo{booktitle}{\emph{Proceedings of the IEEE/CVF
  Conference on Computer Vision and Pattern Recognition}}.
  \bibinfo{pages}{2604--2613}.
\newblock


\bibitem[Liu et~al\mbox{.}(2015)]%
        {liu2015faceattributes}
\bibfield{author}{\bibinfo{person}{Ziwei Liu}, \bibinfo{person}{Ping Luo},
  \bibinfo{person}{Xiaogang Wang}, {and} \bibinfo{person}{Xiaoou Tang}.}
  \bibinfo{year}{2015}\natexlab{}.
\newblock \showarticletitle{Deep Learning Face Attributes in the Wild}. In
  \bibinfo{booktitle}{\emph{Proceedings of International Conference on Computer
  Vision (ICCV)}}.
\newblock


\bibitem[Lopes et~al\mbox{.}(2022)]%
        {lopes2022cross}
\bibfield{author}{\bibinfo{person}{Ivan Lopes}, \bibinfo{person}{Tuan-Hung Vu},
  {and} \bibinfo{person}{Raoul de Charette}.} \bibinfo{year}{2022}\natexlab{}.
\newblock \showarticletitle{Cross-task Attention Mechanism for Dense Multi-task
  Learning}.
\newblock \bibinfo{journal}{\emph{arXiv preprint arXiv:2206.08927}}
  (\bibinfo{year}{2022}).
\newblock


\bibitem[Loshchilov and Hutter(2017)]%
        {loshchilov2017decoupled}
\bibfield{author}{\bibinfo{person}{Ilya Loshchilov} {and}
  \bibinfo{person}{Frank Hutter}.} \bibinfo{year}{2017}\natexlab{}.
\newblock \showarticletitle{Decoupled weight decay regularization}.
\newblock \bibinfo{journal}{\emph{arXiv preprint arXiv:1711.05101}}
  (\bibinfo{year}{2017}).
\newblock


\bibitem[{Microsoft}(2022)]%
        {msweb}
\bibfield{author}{\bibinfo{person}{{Microsoft}}.}
  \bibinfo{year}{2022}\natexlab{}.
\newblock \bibinfo{title}{Microsoft computer vision API.}
\newblock
\newblock
\newblock
\shownote{\url{http://azure.microsoft.com/en-us/products/cognitive-services/computer-vision/}}.


\bibitem[Mironov(2017)]%
        {mironov2017renyi}
\bibfield{author}{\bibinfo{person}{Ilya Mironov}.}
  \bibinfo{year}{2017}\natexlab{}.
\newblock \showarticletitle{R{\'e}nyi differential privacy}. In
  \bibinfo{booktitle}{\emph{2017 IEEE 30th computer security foundations
  symposium (CSF)}}. IEEE, \bibinfo{pages}{263--275}.
\newblock


\bibitem[Nathan~Silberman and Fergus(2012)]%
        {SilbermanECCV12}
\bibfield{author}{\bibinfo{person}{Pushmeet~Kohli Nathan~Silberman,
  Derek~Hoiem} {and} \bibinfo{person}{Rob Fergus}.}
  \bibinfo{year}{2012}\natexlab{}.
\newblock \showarticletitle{Indoor Segmentation and Support Inference from RGBD
  Images}. In \bibinfo{booktitle}{\emph{ECCV}}.
\newblock


\bibitem[Nguyen et~al\mbox{.}(2019)]%
        {nguyen2019multi}
\bibfield{author}{\bibinfo{person}{Huy~H Nguyen}, \bibinfo{person}{Fuming
  Fang}, \bibinfo{person}{Junichi Yamagishi}, {and} \bibinfo{person}{Isao
  Echizen}.} \bibinfo{year}{2019}\natexlab{}.
\newblock \showarticletitle{Multi-task learning for detecting and segmenting
  manipulated facial images and videos}. In \bibinfo{booktitle}{\emph{2019 IEEE
  10th International Conference on Biometrics Theory, Applications and Systems
  (BTAS)}}. IEEE, \bibinfo{pages}{1--8}.
\newblock


\bibitem[Nimi et~al\mbox{.}(2022)]%
        {nimi2022chimera}
\bibfield{author}{\bibinfo{person}{Sumaiya~Tabassum Nimi},
  \bibinfo{person}{Md~Adnan Arefeen}, \bibinfo{person}{Md~Yusuf~Sarwar Uddin},
  \bibinfo{person}{Biplob Debnath}, {and} \bibinfo{person}{Srimat Chakradhar}.}
  \bibinfo{year}{2022}\natexlab{}.
\newblock \showarticletitle{Chimera: Context-aware splittable deep multitasking
  models for edge intelligence}. In \bibinfo{booktitle}{\emph{2022 IEEE
  International Conference on Smart Computing (SMARTCOMP)}}. IEEE,
  \bibinfo{pages}{70--77}.
\newblock


\bibitem[Pan et~al\mbox{.}(2022)]%
        {pan2022integration}
\bibfield{author}{\bibinfo{person}{Xuran Pan}, \bibinfo{person}{Chunjiang Ge},
  \bibinfo{person}{Rui Lu}, \bibinfo{person}{Shiji Song},
  \bibinfo{person}{Guanfu Chen}, \bibinfo{person}{Zeyi Huang}, {and}
  \bibinfo{person}{Gao Huang}.} \bibinfo{year}{2022}\natexlab{}.
\newblock \showarticletitle{On the integration of self-attention and
  convolution}. In \bibinfo{booktitle}{\emph{Proceedings of the IEEE/CVF
  Conference on Computer Vision and Pattern Recognition}}.
  \bibinfo{pages}{815--825}.
\newblock


\bibitem[Seichter et~al\mbox{.}(2021)]%
        {seichter2021efficient}
\bibfield{author}{\bibinfo{person}{Daniel Seichter}, \bibinfo{person}{Mona
  K{\"o}hler}, \bibinfo{person}{Benjamin Lewandowski}, \bibinfo{person}{Tim
  Wengefeld}, {and} \bibinfo{person}{Horst-Michael Gross}.}
  \bibinfo{year}{2021}\natexlab{}.
\newblock \showarticletitle{Efficient rgb-d semantic segmentation for indoor
  scene analysis}. In \bibinfo{booktitle}{\emph{2021 IEEE International
  Conference on Robotics and Automation (ICRA)}}. IEEE,
  \bibinfo{pages}{13525--13531}.
\newblock


\bibitem[Standley et~al\mbox{.}(2020)]%
        {standley2020tasks}
\bibfield{author}{\bibinfo{person}{Trevor Standley}, \bibinfo{person}{Amir
  Zamir}, \bibinfo{person}{Dawn Chen}, \bibinfo{person}{Leonidas Guibas},
  \bibinfo{person}{Jitendra Malik}, {and} \bibinfo{person}{Silvio Savarese}.}
  \bibinfo{year}{2020}\natexlab{}.
\newblock \showarticletitle{Which tasks should be learned together in
  multi-task learning?}. In \bibinfo{booktitle}{\emph{International Conference
  on Machine Learning}}. PMLR, \bibinfo{pages}{9120--9132}.
\newblock


\bibitem[Thapa et~al\mbox{.}(2022)]%
        {thapa2022splitfed}
\bibfield{author}{\bibinfo{person}{Chandra Thapa}, \bibinfo{person}{Pathum
  Chamikara~Mahawaga Arachchige}, \bibinfo{person}{Seyit Camtepe}, {and}
  \bibinfo{person}{Lichao Sun}.} \bibinfo{year}{2022}\natexlab{}.
\newblock \showarticletitle{Splitfed: When federated learning meets split
  learning}. In \bibinfo{booktitle}{\emph{Proceedings of the AAAI Conference on
  Artificial Intelligence}}, Vol.~\bibinfo{volume}{36}.
  \bibinfo{pages}{8485--8493}.
\newblock


\bibitem[Vogels et~al\mbox{.}(2019)]%
        {vogels2019powersgd}
\bibfield{author}{\bibinfo{person}{Thijs Vogels}, \bibinfo{person}{Sai~Praneeth
  Karimireddy}, {and} \bibinfo{person}{Martin Jaggi}.}
  \bibinfo{year}{2019}\natexlab{}.
\newblock \showarticletitle{PowerSGD: Practical low-rank gradient compression
  for distributed optimization}.
\newblock \bibinfo{journal}{\emph{Advances in Neural Information Processing
  Systems}}  \bibinfo{volume}{32} (\bibinfo{year}{2019}).
\newblock


\bibitem[Wang et~al\mbox{.}(2022)]%
        {wang2022cross}
\bibfield{author}{\bibinfo{person}{Li Wang}, \bibinfo{person}{Dong Li},
  \bibinfo{person}{Han Liu}, \bibinfo{person}{Jinzhang Peng},
  \bibinfo{person}{Lu Tian}, {and} \bibinfo{person}{Yi Shan}.}
  \bibinfo{year}{2022}\natexlab{}.
\newblock \showarticletitle{Cross-dataset collaborative learning for semantic
  segmentation in autonomous driving}. In \bibinfo{booktitle}{\emph{Proceedings
  of the AAAI Conference on Artificial Intelligence}},
  Vol.~\bibinfo{volume}{36}. \bibinfo{pages}{2487--2494}.
\newblock


\bibitem[Wang et~al\mbox{.}(2021)]%
        {wang2021deep}
\bibfield{author}{\bibinfo{person}{Mengyang Wang}, \bibinfo{person}{Zhicong
  Zhang}, \bibinfo{person}{Jiahui Li}, \bibinfo{person}{Mengyao Ma}, {and}
  \bibinfo{person}{Xiaopeng Fan}.} \bibinfo{year}{2021}\natexlab{}.
\newblock \showarticletitle{Deep joint source-channel coding for multi-task
  network}.
\newblock \bibinfo{journal}{\emph{IEEE Signal Processing Letters}}
  \bibinfo{volume}{28} (\bibinfo{year}{2021}), \bibinfo{pages}{1973--1977}.
\newblock


\bibitem[Wang et~al\mbox{.}(2020)]%
        {wang2020eca}
\bibfield{author}{\bibinfo{person}{Qilong Wang}, \bibinfo{person}{Banggu Wu},
  \bibinfo{person}{Pengfei Zhu}, \bibinfo{person}{Peihua Li},
  \bibinfo{person}{Wangmeng Zuo}, {and} \bibinfo{person}{Qinghua Hu}.}
  \bibinfo{year}{2020}\natexlab{}.
\newblock \showarticletitle{ECA-Net: Efficient Channel Attention for Deep
  Convolutional Neural Networks}. In \bibinfo{booktitle}{\emph{The IEEE
  Conference on Computer Vision and Pattern Recognition (CVPR)}}.
\newblock


\bibitem[Wang et~al\mbox{.}(2019)]%
        {wang2019lednet}
\bibfield{author}{\bibinfo{person}{Yu Wang}, \bibinfo{person}{Quan Zhou},
  \bibinfo{person}{Jia Liu}, \bibinfo{person}{Jian Xiong},
  \bibinfo{person}{Guangwei Gao}, \bibinfo{person}{Xiaofu Wu}, {and}
  \bibinfo{person}{Longin~Jan Latecki}.} \bibinfo{year}{2019}\natexlab{}.
\newblock \showarticletitle{Lednet: A lightweight encoder-decoder network for
  real-time semantic segmentation}. In \bibinfo{booktitle}{\emph{2019 IEEE
  International Conference on Image Processing (ICIP)}}. IEEE,
  \bibinfo{pages}{1860--1864}.
\newblock


\bibitem[Wang et~al\mbox{.}(2004)]%
        {wang2004image}
\bibfield{author}{\bibinfo{person}{Zhou Wang}, \bibinfo{person}{Alan~C Bovik},
  \bibinfo{person}{Hamid~R Sheikh}, {and} \bibinfo{person}{Eero~P Simoncelli}.}
  \bibinfo{year}{2004}\natexlab{}.
\newblock \showarticletitle{Image quality assessment: from error visibility to
  structural similarity}.
\newblock \bibinfo{journal}{\emph{IEEE transactions on image processing}}
  \bibinfo{volume}{13}, \bibinfo{number}{4} (\bibinfo{year}{2004}),
  \bibinfo{pages}{600--612}.
\newblock


\bibitem[Xiao et~al\mbox{.}(2020)]%
        {xiao2020adversarial}
\bibfield{author}{\bibinfo{person}{Taihong Xiao}, \bibinfo{person}{Yi-Hsuan
  Tsai}, \bibinfo{person}{Kihyuk Sohn}, \bibinfo{person}{Manmohan Chandraker},
  {and} \bibinfo{person}{Ming-Hsuan Yang}.} \bibinfo{year}{2020}\natexlab{}.
\newblock \showarticletitle{Adversarial learning of privacy-preserving and
  task-oriented representations}. In \bibinfo{booktitle}{\emph{Proceedings of
  the AAAI Conference on Artificial Intelligence}}, Vol.~\bibinfo{volume}{34}.
  \bibinfo{pages}{12434--12441}.
\newblock


\bibitem[Yang et~al\mbox{.}(2021)]%
        {yang2021video}
\bibfield{author}{\bibinfo{person}{Wenhan Yang}, \bibinfo{person}{Haofeng
  Huang}, \bibinfo{person}{Yueyu Hu}, \bibinfo{person}{Ling-Yu Duan}, {and}
  \bibinfo{person}{Jiaying Liu}.} \bibinfo{year}{2021}\natexlab{}.
\newblock \showarticletitle{Video Coding for Machine: Compact Visual
  Representation Compression for Intelligent Collaborative Analytics}.
\newblock \bibinfo{journal}{\emph{arXiv preprint arXiv:2110.09241}}
  (\bibinfo{year}{2021}).
\newblock


\bibitem[Yao et~al\mbox{.}(2020)]%
        {yao2020deep}
\bibfield{author}{\bibinfo{person}{Shuochao Yao}, \bibinfo{person}{Jinyang Li},
  \bibinfo{person}{Dongxin Liu}, \bibinfo{person}{Tianshi Wang},
  \bibinfo{person}{Shengzhong Liu}, \bibinfo{person}{Huajie Shao}, {and}
  \bibinfo{person}{Tarek Abdelzaher}.} \bibinfo{year}{2020}\natexlab{}.
\newblock \showarticletitle{Deep compressive offloading: Speeding up neural
  network inference by trading edge computation for network latency}. In
  \bibinfo{booktitle}{\emph{Proceedings of the 18th Conference on Embedded
  Networked Sensor Systems}}. \bibinfo{pages}{476--488}.
\newblock


\bibitem[Yousefpour et~al\mbox{.}(2021)]%
        {yousefpour2021opacus}
\bibfield{author}{\bibinfo{person}{Ashkan Yousefpour}, \bibinfo{person}{Igor
  Shilov}, \bibinfo{person}{Alexandre Sablayrolles}, \bibinfo{person}{Davide
  Testuggine}, \bibinfo{person}{Karthik Prasad}, \bibinfo{person}{Mani Malek},
  \bibinfo{person}{John Nguyen}, \bibinfo{person}{Sayan Ghosh},
  \bibinfo{person}{Akash Bharadwaj}, \bibinfo{person}{Jessica Zhao},
  {et~al\mbox{.}}} \bibinfo{year}{2021}\natexlab{}.
\newblock \showarticletitle{Opacus: User-friendly differential privacy library
  in PyTorch}.
\newblock \bibinfo{journal}{\emph{arXiv preprint arXiv:2109.12298}}
  (\bibinfo{year}{2021}).
\newblock


\bibitem[Zhang et~al\mbox{.}(2022)]%
        {zhang2022resnest}
\bibfield{author}{\bibinfo{person}{Hang Zhang}, \bibinfo{person}{Chongruo Wu},
  \bibinfo{person}{Zhongyue Zhang}, \bibinfo{person}{Yi Zhu},
  \bibinfo{person}{Haibin Lin}, \bibinfo{person}{Zhi Zhang},
  \bibinfo{person}{Yue Sun}, \bibinfo{person}{Tong He}, \bibinfo{person}{Jonas
  Mueller}, \bibinfo{person}{R Manmatha}, {et~al\mbox{.}}}
  \bibinfo{year}{2022}\natexlab{}.
\newblock \showarticletitle{Resnest: Split-attention networks}. In
  \bibinfo{booktitle}{\emph{Proceedings of the IEEE/CVF Conference on Computer
  Vision and Pattern Recognition}}. \bibinfo{pages}{2736--2746}.
\newblock


\bibitem[Zhang and Yang(2021)]%
        {zhang2021sa}
\bibfield{author}{\bibinfo{person}{Qing-Long Zhang} {and}
  \bibinfo{person}{Yu-Bin Yang}.} \bibinfo{year}{2021}\natexlab{}.
\newblock \showarticletitle{Sa-net: Shuffle attention for deep convolutional
  neural networks}. In \bibinfo{booktitle}{\emph{ICASSP 2021-2021 IEEE
  International Conference on Acoustics, Speech and Signal Processing
  (ICASSP)}}. IEEE, \bibinfo{pages}{2235--2239}.
\newblock


\bibitem[Zhang et~al\mbox{.}(2021)]%
        {zhang2021msfc}
\bibfield{author}{\bibinfo{person}{Zhicong Zhang}, \bibinfo{person}{Mengyang
  Wang}, \bibinfo{person}{Mengyao Ma}, \bibinfo{person}{Jiahui Li}, {and}
  \bibinfo{person}{Xiaopeng Fan}.} \bibinfo{year}{2021}\natexlab{}.
\newblock \showarticletitle{Msfc: Deep feature compression in multi-task
  network}. In \bibinfo{booktitle}{\emph{2021 IEEE International Conference on
  Multimedia and Expo (ICME)}}. IEEE, \bibinfo{pages}{1--6}.
\newblock


\end{thebibliography}


\end{document}